\pdfoutput=1
\documentclass[11pt]{article}

\usepackage[]{acl}

\usepackage{times}
\usepackage{latexsym}
\usepackage{kotex}
\usepackage{graphicx}
\usepackage{multirow}
\usepackage{listings} 
\usepackage{tabularx}
\usepackage{booktabs}
\usepackage{xcolor}
\usepackage{colortbl}
\usepackage{hhline}
\usepackage{amsmath}

\newcolumntype{Y}{>{\centering\arraybackslash}X}

\newcommand{\sysname}{CS1QA}

\usepackage[T1]{fontenc}
\usepackage[utf8]{inputenc}

\usepackage{microtype}

\title{\sysname{}: A Dataset for Assisting Code-based Question Answering in an Introductory Programming Course}

\author{Changyoon Lee, Yeon Seonwoo, Alice Oh \\
    School of Computing, KAIST \\
  \texttt{changyoon.lee@kaist.ac.kr, yeon.seonwoo@kaist.ac.kr} \\
  \texttt{alice.oh@kaist.edu}
  }

\begin{document}
\maketitle
\begin{abstract}
We introduce~\sysname{}, a dataset for code-based question answering in the programming education domain. 
~\sysname{} consists of 9,237 question-answer pairs gathered from chat logs in an introductory programming class using Python, and 17,698 unannotated chat data with code\footnote{The code and the data used in this paper can be found at \url{https://github.com/cyoon47/CS1QA}.}. 
Each question is accompanied with the student's code, and the portion of the code relevant to answering the question. 
We carefully design the annotation process to construct~\sysname{}, and analyze the collected dataset in detail.
The tasks for~\sysname{} are to predict the question type, the relevant code snippet given the question and the code and retrieving an answer from the annotated corpus.
Results for the experiments on several baseline models are reported and thoroughly analyzed. The tasks for~\sysname{} challenge models to understand both the code and natural language. This unique dataset can be used as a benchmark for source code comprehension and question answering in the educational setting.

\end{abstract}

\section{Introduction}
% Datasets - attempts at question answering using natural language
Question answering (QA) studies systems that understand questions and the relevant context to provide answers. 
% The questions, their context and input modes vary according to the domain. 
Question forms include single document QA~\cite{rajpurkar-etal-2016-squad}, multi-hop QA~\cite{yang2018hotpotqa}, conversational QA~\cite{reddy-etal-2019-coqa}, and open domain QA~\cite{kwiatkowski2019natural}. Questions about specific domains are asked in NewsQA~\cite{trischler2016newsqa} and TechQA~\cite{castelli-etal-2020-techqa}, and images are provided with the question in visual QA~\cite{antol2015vqa}. Another interesting field of QA asks questions about source code
%, which is fundamentally different from natural language
~\cite{liu2021codeqa}.

% Education domain + code 
A useful application of QA is in the educational domain. Asking questions and getting the answer is an essential and efficient means of learning. 
In this paper, we focus on QA for programming education, where both the input modes and the domain pose interesting challenges.
Answering these questions requires reading and understanding both source code and natural language questions. In addition, students' questions are often complex, demanding thorough understanding of the context such as the intention and the educational goal to answer them.

\begin{figure}[t]
    \centering
    \includegraphics[width=0.8\linewidth]{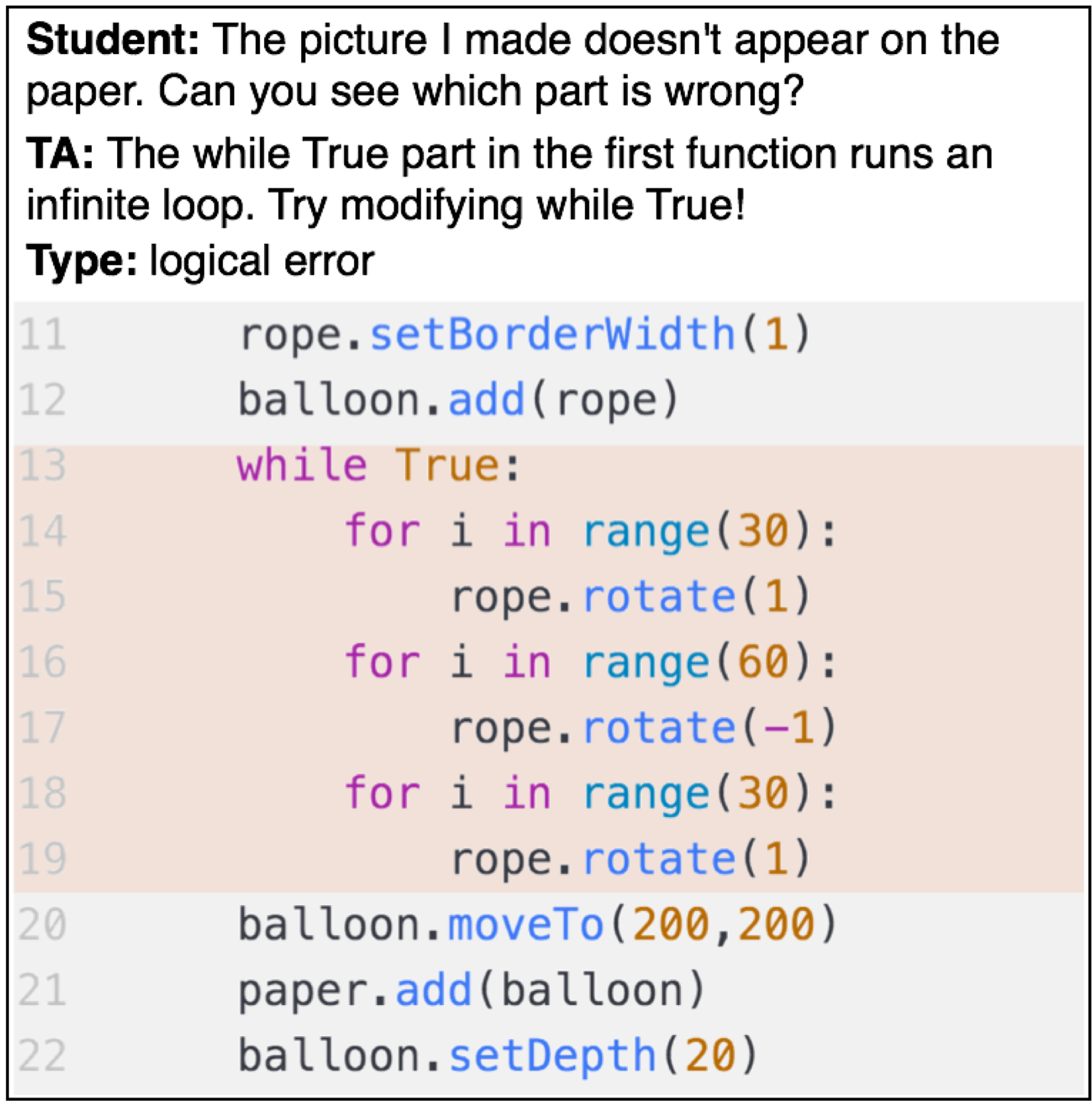}
    \caption{An example of our data tuple. Each data tuple consists of \{question, answer, question type, code, relevant code lines\}.
    %In this example, the student asks for help to resolve the logical error in their code (grey). The TA replies to the question.
    We annotate the type of each question and the code lines (orange) relevant to the question.}
    \label{fig:example}
\end{figure}

% \ysw{두번째 paragraph는 최근에 나온모델에 대한 제한점이 간단하게 들어가면 좋을거같아요.}
%\ysw{두번째 paragraph를 조금 바꿔봤어요.}
Recently, models that understand programming languages (PL) have been studied, and show promising results in diverse code comprehension tasks~\cite{alon2018code2seq,feng2020codebert, guo2021graphcodebert}. 
% Datasets based on open source projects on Github are used to train these models, such as Code2Seq
% \cite{alon2018code2seq}, CodeBERT \cite{feng2020codebert}, and GraphCodeBERT \cite{guo2021graphcodebert}.
However, these models have limitations to support question answering.
They are not trained on datasets containing questions about the code and are not designed for QA tasks. Also, many assume fully functional code as input, while students' code contains diverse syntax and logical errors and is often incomplete.

% \ysw{Also, questions and code descriptions are different in terms of the questioner are inapprehensive to their codes.}

% Several models such as Code2Seq~\cite{alon2018code2seq}, CodeBERT~\cite{feng2020codebert}, and GraphCodeBERT~\cite{guo2021graphcodebert} have been evaluated on tasks such as code document generation and code search. However, many of these models cannot take codes with errors as inputs, and their performance on code-based QA is unknown, due to the limited number of code-based datasets and tasks. 

% Models' understanding of natural language questions has shown tremendous improvements with the abundance of large scale data for training question answering (QA) models.
% Numerous QA datasets have introduced tasks that challenge QA models to adopt different strategies to understand the text. Questions from these datasets require QA models to provide answers by reading short passages, multiple linked passages or conversations in addition to the passage~\cite{rajpurkar-etal-2016-squad, yang-etal-2018-hotpotqa, reddy-etal-2019-coqa}. The diversity and availability of the datasets have driven the recent progress in QA.

To address this issue, we introduce \sysname{}, a new dataset with tasks for code-based question answering in programming education.
%The data was collected mostly in Korean and then translated into English for use in models pretrained in English. \ysw{이 문장은 나중에 설명하는게 좋을거같아요.}
%\sysname{} aims to serve as a dataset for a new type of question answering based on code. 
% The code in \sysname{} serves a similar purpose as the passages in other QA datasets. 
% However, code has unique characteristics not present in natural languages. A small change in structure, such as the whitespace, can completely change the meaning of the code, while it is less influential for natural language. Also, vocabulary of code is not usually shared, as it is often uniquely defined within the documents.
% %\highlight{However, while in other QA datasets, the passages explicitly contain the answers for the questions, the code in \sysname{} usually does not. The code provides only the information needed to answer the question, and further reasoning is required to actually provide an answer.} 
% These differences introduce new challenges for natural language models when answering questions in \sysname{}.
%Programming languages have many differences from natural languages. While both of them convey some meaning in the form of text, programming languages represent instructions with specific objectives. We believe that this fundamental difference will introduce difficulties for natural language models in understanding code.
% The data for \sysname{} are collected from  in an online programming course via crowdsourcing. 
Questions and answers about programming are collected from the naturally occurring chat messages between students and TAs. The question type and the code snippet relevant to answering the question
%, which provide useful information for answering, 
are also collected. The final \sysname{} dataset consists of question, question type, answer, and code annotated with relevant lines.
The data is collected mostly in Korean and then machine-translated into English and quality-checked for easy application on models pretrained in English. Figure~\ref{fig:example} shows an example of our data. We also include two-semesters' worth of TA-student chat log data consisting of 17,698 chat sessions and the corresponding code.

We design three tasks for the \sysname{} dataset. Type classification task asks the model to predict the question type. Code line selection task asks the model to select lines of code that are relevant to answering the given question. Answer retrieval task finds a similar question already answered, and uses its answer as the answer to the given question.
The outputs for these tasks can help the students debug their code and the TAs spend less time and effort when answering the students' questions.

Finally, we implement and test baseline models, RoBERTa~\cite{liu2019roberta}, CodeBERT~\cite{feng2020codebert} and XLM-RoBERTa~\cite{conneau2020unsupervised}, on the type classification and code line selection tasks. The finetuned models achieve accuracies up to 76.65\% for the type classification task. The relatively low F1 scores of 57.57\% for the line selection task suggest that the task is challenging for current language models. We use DPR~\cite{karpukhin-etal-2020-dense} to retrieve the most similar question and its answer. We compare the retrieved answer with the gold label answer, and achieve a BLEU-1 score of 13.07, which shows incompetent performance of answer retrieval on~\sysname{} dataset. We show with a qualitative evaluation the model behavior with different inputs for the first two tasks.
%, possibly due to poor understanding of the code.
Our contributions are as follows:
\begin{itemize}
    \item We present \sysname{}, a dataset containing 9,237 question-answer-code triples from a programming course, annotated with question types and relevant code lines. The dataset's contribution includes student-TA chat logs in a live classroom.
    \item We introduce three tasks, question type classification, code line selection and answer retrieval, that require models to comprehend the text and provide useful output for TAs and students when answering questions.
    \item We present the results of baseline models on the tasks. Models find the tasks in \sysname{} challenging, and have much room for improvement in performance.
\end{itemize}

% Programming education / code understanding(?) - current status 

% \sysname{} / experiment/results

\section{Related Work}
\paragraph{Code-based Datasets}
Recently, research dealing with large amounts of source code data has gained attention. Often, the source code data is collected ad hoc for the purpose of the research \cite{allamanis2018learning, brockschmidt2018generative, clement2020pymt5}. 
Several datasets have been released to aid research in source code comprehension, and avoid repeated crawling and processing of source code data. These datasets serve as benchmarks for different tasks that test the ability to understand code. Such datasets include:
ETH Py150 corpus~\cite{raychev2016probabilistic}, CodeNN~\cite{iyer2016summarizing}, CodeSearchNet~\cite{husain2020codesearchnet} and CodeQA~\cite{liu2021codeqa}. 
We compare these datasets with~\sysname{} in Table~\ref{tab:code_datasets}.

\begin{table*}[h]
\centering
\resizebox{\textwidth}{!}{%
\begin{tabular}{@{}cclcc@{}}
\toprule
Dataset       & Programming Language                    & Data Format                                    & Dataset Size        & Data Source          \\ \midrule
ETH Py150     & Python                                  & Parsed AST                                     & 7.4M files          & GitHub               \\
\rowcolor[HTML]{DDDDDD} 
CodeNN        & C\#, SQL                                & {\color[HTML]{000000} Title, question, answer} & $\sim$187,000 pairs & StackOverflow        \\
CodeSearchNet & Go, Java, JavaScript, PHP, Python, Ruby & Comment, code                                  & $\sim$2M pairs      & GitHub               \\
\rowcolor[HTML]{DDDDDD} 
CodeQA        & Java, Python                            & Question, answer, code                         & $\sim$190,000 pairs & GitHub               \\ \midrule
CS1QA         & Python                                  & Chat log, question, answer, type, code         & 9,237 pairs         & Real-world classroom \\ \bottomrule
\end{tabular}%
}
\caption{Comparison between different code-based datasets and~\sysname{}.}
\label{tab:code_datasets}
\end{table*}

In an educational setting, students' code presents different chracteristics from code in these datasets: 1) students' code is often incomplete, 2) there are many errors in the code, 3) students' code is generally longer than code used in existing datasets, and 4) questions and answers from students and TAs provide important additional information. In~\sysname{}, we present a dataset more suited for the programming education context.

\paragraph{Source Code Comprehension}
In the domain of machine learning and software engineering, understanding and representing source code using neural networks has become an important approach. Different approaches make use of different characteristics present in programming languages. One such characteristic is the rich syntactic information found in the source code's abstract syntax tree (AST). Code2seq~\cite{alon2018code2seq} passes paths in the AST through an encoder-decoder network to represent code. The graph structure of AST has been exploited in other research for source code representation on downstream tasks such as variable misuse detection, code generation, natural language code search and program repair~\cite{allamanis2018learning,brockschmidt2018generative,guo2021graphcodebert, yasunaga2020graph}.
Source code text itself is used in models such as CodeBERT~\cite{feng2020codebert}, CuBERT~\cite{kanade2020learning} and DeepFix~\cite{gupta2017deepfix} for use in tasks such as natural language code search, finding function-docstring mismatch and program repair.

The tasks that these methods are trained on target expert software engineers and programmers who can gain significant benefit with support by the model. On the other hand, students learning programming have different objectives and require fitting support by the models. Rather than getting an answer quickly, students seek to 
Students ask lots of questions while learning, and thus question answering for code is needed. 
\sysname{} focuses on code-based question answering and can be used as training data and a benchmark for neural models in an education setting. The~\sysname{} data can also be used for other tasks than QA, such as program repair and code search.

\section{\sysname{} Dataset}
\subsection{Data Source}
The data for~\sysname{} is collected from an introductory programming course conducted online\footnote{Elice \url{https://elice.io/}}. Students complete lab sessions consisting of several programming tasks and students and TAs ask questions to each other using a synchronous chat feature. We make use of the chat logs as the source for the natural question and the corresponding answer. These chat logs are either in Korean or in English.
The student's code history is also stored for each programming task for every keystroke the student makes. This allows us to extract the code status at the exact time the question is asked, which provides valuable context for the question. We take this code as the context for the given question. 
The thorough code history and the student-TA chat logs are a unique and important contribution of~\sysname{}. \sysname{} also contributes with data from multiple students working on the same set of problems.

\subsection{Question Type Categorization}
Answering different types of questions requires understanding the different intentions and information - answering questions about errors requires identifying the erroneous code and answering questions about algorithms requires understanding the overall program flow. As the different question types affect the answering approach and location of code to look at, knowing them in advance can be beneficial in the QA and code selection tasks. 

\citet{allamanis2013and} have categorized questions asking for help in coding on Stack Overflow into five types. We adapt these types to students' questions. In addition, we define the ``Task'' type that asks about the requirements of the task. TAs’ question types are derived from the official instructions by the course instructors given in the beginning of the semester. TAs were instructed to ask questions that gauge students' understanding of their implementation, for example by asking the meaning of the code and reasoning behind the implementation. TAs' probing questions are categorized into five types: \textit{Comparison}, \textit{Reasoning}, \textit{Explanation}, \textit{Meaning}, and \textit{Guiding}.
Examples for the question types can be found in Table~\ref{tab:question_examples}. We present intentions of the question types in Table~\ref{tab:question_types}. 
\begin{table*}[h]
\centering
\scriptsize
\begin{tabularx}{\textwidth}{ccXX}
\toprule
Question Type                                                   & Allamanis' Type         & \multicolumn{1}{c}{Question}                                                                                                                                      & \multicolumn{1}{c}{Answer}                                                                                                                     \\ \midrule
\rowcolor[HTML]{DDDDDD} 
\begin{tabular}[c]{@{}c@{}}Code\\ Understanding\end{tabular}    & How/why something works & Why is the print cards function at the bottom of the check function? Can I not have it?                                                                           & This is because if two cards match through a check, you have to show them two cards.                                                           \\
Logical Error                                                   & Do not work             & Now, the file is created, but when I go inside and look at the value, it seems to be a little different from the one requested in the problem. & You seem to have forgotten the line break in the middle I think you can add \textbackslash{}n                                                  \\
\rowcolor[HTML]{DDDDDD} 
Error                                                           & Do not work             & I don't know where the task 2 error came from....                                                                                                                 & When creating image There is a negative number in the image size Please do something like absolute value                                       \\
\begin{tabular}[c]{@{}c@{}}Function/Syntax\\ Usage\end{tabular} & Way of using            & And I forgot how to make a blank image                                                                                                                            & Blank images can be created with new\_img= create\_picture(width,height)!                                                                      \\
\rowcolor[HTML]{DDDDDD} 
Algorithm                                                       & How to implement        & So, what if there is any other way to count the number including the number not included in the randomly created list?                                            & Parameter: a list which is returned from drawing\_integers Count integers function is supposed to take that as input                           \\
Task                                                            & -                       & Isn't it a task in which the number of iteration steps changes according to the input value?                                                                      & Yes, but the value of x must also change according to the input value!                                                                         \\ \midrule
\rowcolor[HTML]{DDDDDD} 
Comparison                                                      & -                       & How was the method of reading and writing different in task1?                                                                                                     & There was a difference between read mode and write mode, open(file name, r) and open (file name, w)                                            \\
Reasoning                                                       & -                       & I've read the task1 code. What is the intention of using continue on line 55?                                                                                     & This is to go back to the beginning without executing the next print function.                                                                 \\
\rowcolor[HTML]{DDDDDD} 
Explanation                                                     & -                       & How do you create new\_img when 'horizontal' is input as Direction in Task2?                                                                                      & I did it like I did with vertical, but since the y value is changing, when I change the y value and run the loop, I did x first among x and y. \\
Meaning                                                         & -                       & Can you explain the role of the global keyword in Task 1?                                                                                                         & If you use a variable specified only within a function, it cannot be used in other functions, so I used it for global variable processing!     \\
\rowcolor[HTML]{DDDDDD} 
Guiding                                                         & -                       & Is there a simpler way to change average\_integers using a function already defined in the python list??                                                          & In average\_integers, it would be more convenient to use the len function when counting the total number of elements. \\ \bottomrule                     
\end{tabularx}%
\caption{Examples of translated and untranslated question and answer texts for each question type in \sysname{}. First column shows our type classification, and second column shows the classification by~\citet{allamanis2013and}. The first six rows in the top part are student question types, the last five rows in the bottom part are TA's probing question types.}
\label{tab:question_examples}
\vspace{-0.3em}
\end{table*}

\begin{table*}[]
\centering
\small
\begin{tabular}{lllcccc}
\toprule
                    & \multicolumn{1}{c}{Q Type}                                       & \multicolumn{1}{c}{Intention}                                                                                                  & \# Q                        & \# Code                     & NA (\%)                      & Span (\%)                    \\ \midrule
                    & \cellcolor[HTML]{DDDDDD}Code Understanding                       & \cellcolor[HTML]{DDDDDD}Understanding the functionality of the code                                                            & \cellcolor[HTML]{DDDDDD}105 & \cellcolor[HTML]{DDDDDD}209  & \cellcolor[HTML]{DDDDDD}33.9 & \cellcolor[HTML]{DDDDDD}11.0 \\
                    & Logical Error                                                    & \begin{tabular}[c]{@{}l@{}}Investigating the cause of\\ the unexpected outputs of the code\end{tabular}                        & 541                         & 1060                         & 16.7                         & 21.6                         \\
                    & \cellcolor[HTML]{DDDDDD}Error                                    & \cellcolor[HTML]{DDDDDD}Resolving syntax errors and exceptions                                                                 & \cellcolor[HTML]{DDDDDD}488 & \cellcolor[HTML]{DDDDDD}959  & \cellcolor[HTML]{DDDDDD}10.1 & \cellcolor[HTML]{DDDDDD}13.0 \\
                    & \begin{tabular}[c]{@{}l@{}}Function/\\ Syntax Usage\end{tabular} & \begin{tabular}[c]{@{}l@{}}Learning correct usage of\\ a function or syntax\end{tabular}                                       & 411                         & 811                          & 55.4                         & 11.8                         \\
                    & \cellcolor[HTML]{DDDDDD}Algorithm                                & \cellcolor[HTML]{DDDDDD}Learning the underlying algorithm for the task                                                         & \cellcolor[HTML]{DDDDDD}603 & \cellcolor[HTML]{DDDDDD}1194 & \cellcolor[HTML]{DDDDDD}50.0 & \cellcolor[HTML]{DDDDDD}19.3 \\
\multirow{-10}{*}{S} & Task                                                             & \begin{tabular}[c]{@{}l@{}}Confirming the goal and requirement\\ of the task\end{tabular}                                      & 677                         & 1322                         & 82.3                         & 15.3                         \\ \midrule
                    & \cellcolor[HTML]{DDDDDD}Reasoning                                & \cellcolor[HTML]{DDDDDD}\begin{tabular}[c]{@{}l@{}}Understanding the reasoning behind\\ student's implementation\end{tabular}  & \cellcolor[HTML]{DDDDDD}402 & \cellcolor[HTML]{DDDDDD}799  & \cellcolor[HTML]{DDDDDD}5.6  & \cellcolor[HTML]{DDDDDD}21.8 \\
                    & \cellcolor[HTML]{FFFFFF}Explanation                              & \cellcolor[HTML]{FFFFFF}\begin{tabular}[c]{@{}l@{}}Checking the validity of\\ student's explanation of their code\end{tabular} & \cellcolor[HTML]{FFFFFF}472 & \cellcolor[HTML]{FFFFFF}940  & \cellcolor[HTML]{FFFFFF}2.6  & \cellcolor[HTML]{FFFFFF}42.8 \\
\multirow{-5}{*}{T} & \cellcolor[HTML]{DDDDDD}Meaning                                  & \cellcolor[HTML]{DDDDDD}\begin{tabular}[c]{@{}l@{}}Checking the meaning of\\ a function or a variable in the code\end{tabular} & \cellcolor[HTML]{DDDDDD}978 & \cellcolor[HTML]{DDDDDD}1943 & \cellcolor[HTML]{DDDDDD}2.1  & \cellcolor[HTML]{DDDDDD}24.2 \\ \bottomrule
\end{tabular}
\caption{Types of questions asked by students (S) and TAs (T) in a programming class. Questions are categorized by different intention and information required to answer them. The number of questions and code snippets collected from the annotators, percentage of \textit{Not applicable} code selections and selected code lines are reported. 
}
\label{tab:question_types}
\end{table*}

\subsection{Collecting Question-Answer Pairs with Question Types}
% Students and TAs communicate synchronously through chat messages available on the course website during the lab sessions.
% We collect the naturally occurring questions and answers from the chat logs between the students and the TAs.

We collected a total of 5,565 chat logs over the course of one semester from 474 students and 47 TAs. After removing the logs where the TA did not participate in the chat, 4,883 chat logs remained. 

We employed crowdworkers with self-reported skill in Python of three or higher on a 5-point Likert Scale to collect the questions. 
%Each crowdworker marked the questions and answers in the chat logs, and selected the corresponding question types. 
Each worker first selected messages in the chat log corresponding to the question and the answer, then selected the question type. Workers were provided with descriptions of the question types with examples before working on the task. Workers were asked to divide the message into individual questions when there were multiple questions or answers in the message. They were instructed to only choose programming related questions, for which the answer is obvious in the chat from the question alone. This ensures that the questions and answers are independent from the chat history. Every chat log was annotated by two workers to ensure the quality of annotation. A total of 20,403 question-answer pairs were collected. 

% but leave the answer generation task as future work, which can be easily designed with \sysname{}. Generating answers from scratch requires problem solving abilities in addition to understanding the code. However, current language models are incapable of the high-level reasoning necessary to generate the answers. In this paper, we focus on evaluating the models on the two tasks that generate supporting evidence for the answer. 

The question and answer texts are machine-translated using Google's Neural Machine Translation model \cite{wu2016gnmt} from Korean to English to form the dataset in two languages. The translation allows for easy integration of \sysname{} data to models pretrained in English, which make up a huge portion of NLP models.

\subsection{Selecting Code Lines}
% The chat logs from which the questions are collected are linked with a lab session in the course. For every lab session, there are up to five tasks that the students have to solve, each with a code file that the students program on.

Providing relevant code snippets allows the answerer to identify the problem more quickly and easily. We annotate the lines of code that provide information necessary to answer the question for use in the code line selection task.

We collected code for all questions asked by students. For the TA probing questions, we collected code for all \textit{Reasoning} and \textit{Meaning} types, and 472 randomly selected \textit{Explanation} questions, for a total of 4677 questions. This keeps a balance in the number of questions for each type. 
\textit{Comparison} questions were left out as they require comparison of code across different tasks, making the annotation and the tasks too complicated. We exclude \textit{Guiding} questions as answering them requires more than just understanding code; the answer is often new algorithms not based on the current code.

We employed crowdworkers who have worked as TAs for the programming course to select the code that the questions refer to. We provided the workers with the collected questions, answers, and the student's code for each task at the time the question was asked. The workers selected the code file for the question and the relevant code lines to answer the question. When reading the code was not necessary to answer the question, the workers were asked to choose \textit{Not Applicable (NA)} for the code selection. For every question, two workers made code annotations. 

% For the purpose of this paper, we treat the code span as the prediction goal for the models, similar to the answer spans in other QA datasets. The answer text was provided to the workers to help them select the code span more accurately. We include the answer text in the final dataset to provide the full context for the questions and for future work of answer generation.

A total of 9,359 code selections were made by the workers. Some of the selections with empty code or incorrectly extracted code were removed from the dataset. The remaining 9,237 questions annotated with type, lab and task numbers, code, code lines and answer make up the final \sysname{} dataset. Every code selection made by the workers is used as gold labels even if the two workers choose different lines. Thus, every question can have up to two correct code selections in different parts of code. An example of the data is found in Appendix~\ref{appendix:example}.

\subsection{Quality Control and Validation}
As the workers worked independently, there were some differences in the annotated data even when they correspond to the same question. There were some questions that were selected by only one worker as well. These questions are further reviewed to ensure the quality of the collected dataset.

Out of 20,403 collected questions, 3,556 questions were selected by only one worker, and 4,787 pairs of questions had some differences between the workers' selections. The remaining questions had perfect agreement between the workers.
The authors reviewed questions selected by only one worker, and those without perfect agreement. Unnecessary words present in only one text were removed and crucial words missing in the question were added to the text while preserving the meaning to make the two texts equal. The conflicts in question types were resolved with the authors' additional vote that made a clear majority in the type selection.

We calculate the inter-rater reliability score with Cohen's Kappa~\cite{cohen1960coefficient} for the question type selection. The Kappa value is calculated between every pair of workers who selected the same question-answer pairs. The mean of the Kappa values is 0.657, which suggests substantial agreement for type classification between the annotators.
% We collect the codes for all edited and reviewed questions and answers, together with those with perfect agreement.

Out of 9,237 questions with code line selection, 2,197 pairs had perfect agreement (100\% overlap), while 1,225 pairs had 0\% overlap. We compute the mean line F1 as the measure for agreement of spans, considering one annotator’s span selection as the ground truth and the other annotator’s selection as prediction. The resulting F1 score is 0.6482. The disagreements are largely due to selecting different but relevant code and selecting different amounts of surrounding context in the code.

\section{Task Definition}
We design three tasks for the \sysname{} dataset that identify important information that leads to the answer. 

The type classification task is to predict the question's type. We use nine types of questions that we categorized as the candidates for classification, each question belonging to a single type. We use the accuracy and macro F1 score as the measure of performance.
The code line selection task is to select lines of code that give relevant information to answer the question. 
The code is a strong supporting context to answering the given question, and this task tests the model's ability to retrieve this critical information.
%\highlight{This task is analogous to the long answer selection in Natural Questions~\cite{kwiatkowski2019natural} and supporting fact selection task in HotpotQA~\cite{yang2018hotpotqa}. - maybe more of selecting a strong supporting context for answering}

For the code line selection task, we use the Exact Match (EM) and line F1 score as the measure of performance, same as the metrics used for supporting fact selection task in HotpotQA~\cite{yang2018hotpotqa}. The EM score measures the proportion of selections that exactly match the ground truth. The line F1 score measures the average overlap between the selected lines and the ground truth selections. The score is computed by treating the selections as bags of lines and calculating their F1 with the annotated lines.
These two tasks take as inputs the lab and task numbers of the question, the questioner (student or TA), question and the code texts.

The answer retrieval task retrieves a similar question given an unseen question, and uses the retrieved question's answer as the answer to the unseen question. BLEU score is calculated between the retrieved answer and the gold label answer. 

Answer generation task given the question and the code context is possible with~\sysname{} dataset. However, meaningfully generating the answer demands a model that understands long and erroneous code, and the natural language question. This poses a significant challenge, and we leave the generation task as future work. 

\section{Dataset Analysis}
% To provide better understanding of the \sysname{} dataset, we analyze the translation quality, text lengths of question, answer and code, the type distribution of the data, and code line selection. 

644 out of 9,237 questions are originally asked in English, while the rest are asked in Korean.
The \sysname{} dataset is split into train, development and test sets in the ratio of 0.6, 0.2 and 0.2 respectively, keeping the ratio of question types in each set the same to ensure equal distribution in all three sets.

\subsection{Text Lengths}
Table~\ref{tab:length_stats} shows the statistics of question and answer token lengths, for data translated to English (EN) and the original (ORIG) data, and the number of lines of code. 
% We observe that the codes are significantly longer in length than questions and answers. 
%\highlight{When separated by whitespace, the mean token lengths of questions and answers translated to English are 15.7 and 27.2 respectively, while the mean token lengths of untranslated questions and answers are 10.9 and 17.6 respectively. The mean token length of codes in \sysname{} is 197.1. The mean number of lines of code in \sysname{} is 76.0.}

% \begin{table}[h]
% \centering
% \small
% \begin{tabular}{c|c|c|c|c|c}
% % \multicolumn{2}{l}{} & \multicolumn{3}{c}{Length of data} \\
%  & Data & Min & Max & Mean & Median \\ \hline
% \multirow{2}{*}{Question} & EN & 1 & 119 & 15.7 & 13 \\
%  & ORIG & 1 & 79 & 10.9 &  9\\ \hline
% \multirow{2}{*}{Answer} & EN & 1 & 272 & 27.2 & 22 \\
%  & ORIG & 1 & 166 & 17.6 & 14\\ \hline
% Code & - & 1 & 655 & 76.0 & 52\\ \hline
% \end{tabular}
% \caption{Statistics of question, answer lengths in tokens and code length in number of lines in \sysname{}.}
% \label{tab:length_stats}
% \vspace{-1em}
% \end{table}

\begin{table}[h]
\centering
\small
\begin{tabular}{@{}c|c|c|c|c|c@{}}
\toprule
\rowcolor[HTML]{FFFFFF} 
                                                   & Data & \multicolumn{1}{l|}{\cellcolor[HTML]{FFFFFF}Min} & Max & Mean & \multicolumn{1}{l}{\cellcolor[HTML]{FFFFFF}Median} \\ \midrule
\rowcolor[HTML]{FFFFFF} 
\cellcolor[HTML]{FFFFFF}                           & EN   & 1                                                & 119 & 15.7 & 13                                                 \\
\rowcolor[HTML]{FFFFFF} 
\multirow{-2}{*}{\cellcolor[HTML]{FFFFFF}Question} & ORIG & {\color[HTML]{000000} 1}                         & 79  & 10.9 & 9                                                  \\ \midrule
\rowcolor[HTML]{FFFFFF} 
\cellcolor[HTML]{FFFFFF}                           & EN   & 1                                                & 272 & 27.2 & 22                                                 \\
\rowcolor[HTML]{FFFFFF} 
\multirow{-2}{*}{\cellcolor[HTML]{FFFFFF}Answer}   & ORIG & 1                                                & 166 & 17.6 & 14                                                 \\ \midrule
\rowcolor[HTML]{FFFFFF} 
Code                                               & -    & 1                                                & 655 & 76.0 & 52                                                 \\ \bottomrule
\end{tabular}
\caption{Statistics of question, answer lengths in tokens and code length in number of lines in \sysname{}.}
\label{tab:length_stats}
\end{table}

%Figure \ref{fig:question_length_distribution} shows that 
The lengths of questions and answers lie mostly between 10 to 30 tokens.
The distributions show long tails for both questions and answers, but answers are more evenly distributed. 
The distribution of token lengths for questions and answers can be found in Appendix~\ref{appendix:distribution}.
%The token length for codes shows a peak between 50 and 100, as shown in Figure \ref{fig:code_length_distribution}. Codes, unlike questions and answers have a wider distribution and do not show an obviously decreasing trend. This trend can be the result of varying difficulties of tasks in the course, with more difficult tasks requiring longer codes to solve.

\begin{figure}[]
\centering
\includegraphics[width=0.42\textwidth]{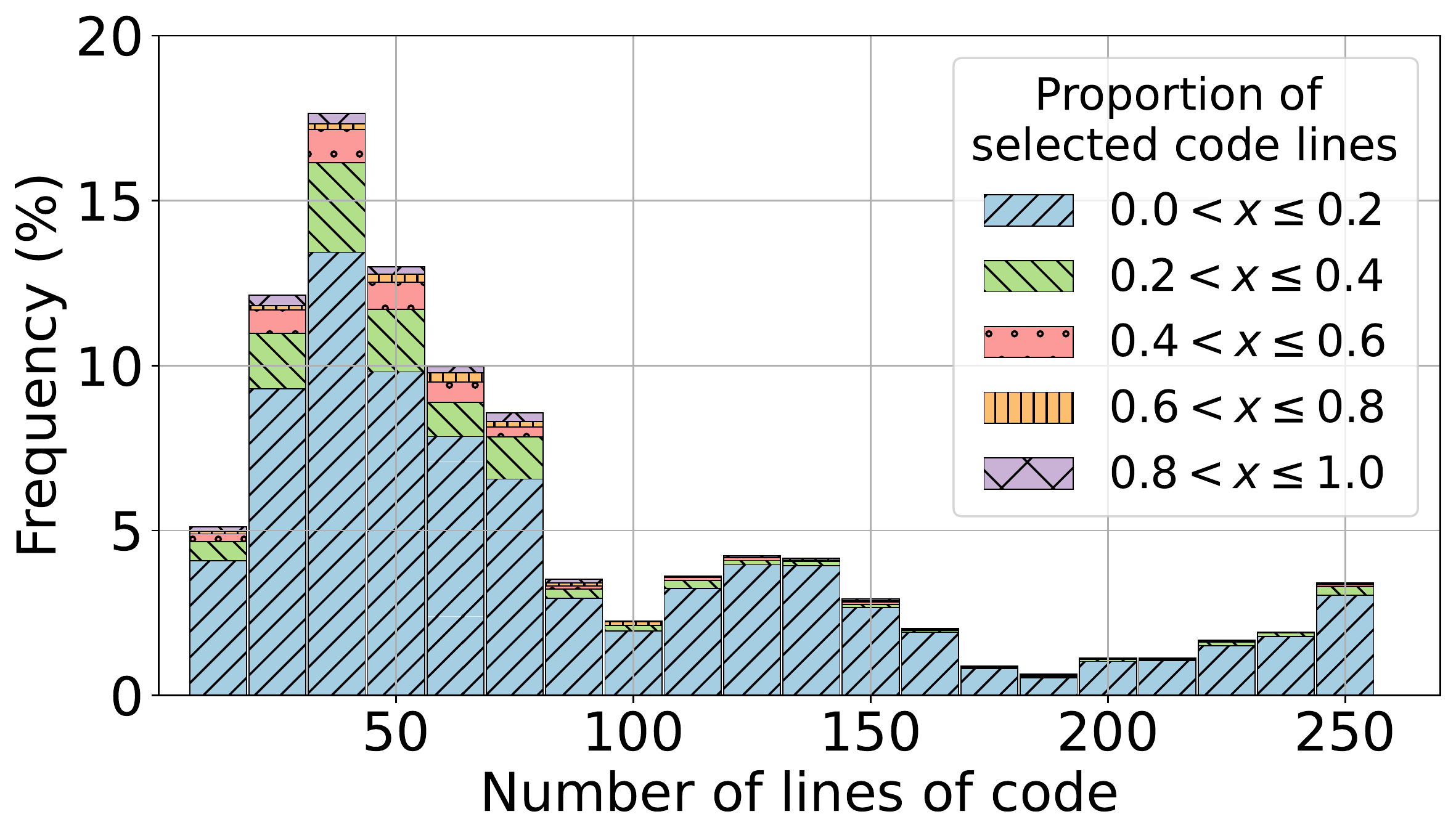}
\caption{The total number of lines in code and the proportion of selected code lines in code (color coded). The last bin contains all code longer than 250 lines.}
\label{fig:line_proportion_v2}
\end{figure}

The number of lines of code shows a peak between 12.5 and 50, as shown in Figure~\ref{fig:line_proportion_v2}. Code snippets have a wider distribution in length. This can be the result of varying difficulties of tasks, with more difficult tasks requiring longer code snippets to solve. The number of lines of code in~\sysname{} is larger than those in other code-based datasets, which can present interesting challenges. 

% \begin{figure*}[h]
% \centering
% \includegraphics[width=0.91\textwidth]{fig/lines_proportion.pdf}
% \caption{The total number of lines in code and the proportion of selected code lines in code (color coded). The last bin contains all codes longer than 250 lines.}
% \label{fig:line_proportion}
% \end{figure*}

% \begin{figure}[h]
% \centering
% \includegraphics[width=0.48\textwidth]{fig/span_length_distribution.eps}
% \caption{The distribution of number of lines selected in code spans. The last bin contains all selections with more than 80 lines of code.}
% \label{fig:code_span_distribution}
% \end{figure}

% \begin{figure}[h]
% \centering
% \includegraphics[width=0.48\textwidth]{fig/question_length_distribution.eps}
% \caption{The distribution of question lengths in number of white space separated tokens. The last bin contains all questions longer than 80 tokens.}
% \label{fig:question_length_distribution}
% \end{figure}

% \begin{figure}[h]
% \centering
% \includegraphics[width=0.48\textwidth]{fig/answer_length_distribution.eps}
% \caption{The distribution of answer lengths in number of white space separated tokens. The last bin contains all answers longer than 100 tokens.}
% \label{fig:answer_length_distribution}
% \end{figure}

% \begin{figure}[h]
% \centering
% \includegraphics[width=0.48\textwidth]{fig/code_length_distribution.eps}
% \caption{The distribution of code lengths in number of white space separated tokens. The last bin contains all codes longer than 1000 tokens.}
% \label{fig:code_length_distribution}
% \end{figure}

% \begin{figure}[h]
% \centering
% \includegraphics[width=0.48\textwidth]{fig/code_line_distribution.eps}
% \caption{The distribution of code lengths in number of lines. The last bin contains all codes longer than 250 lines.}
% \label{fig:code_line_distribution}
% \end{figure}

\subsection{Question Type Distribution}
We present the distribution of question types in Table~\ref{tab:question_types} in number of questions collected and number of code snippets collected. 

The \sysname{} dataset contains similar number of questions for each student type of questions, except for \textit{Code Understanding} type, which contains significantly fewer questions. One plausible reason for this is that most of the tasks require writing the program from scratch, thus students ask fewer questions about the skeleton code.

There are more \textit{Meaning} questions than other types of TAs' probing questions. This can be because TAs often ask the students about the meanings of functions and variables to make sure that the students understand the code they wrote for each task.

\subsection{Code Line Distribution}
The average number of selected code lines is 13.0. A majority of the questions can be answered by looking at fewer than 20 lines of code. The number of selected code lines can be a gauge of the difficulty of answering the questions; a longer selection means that one has to read and understand a larger amount of code. The detailed distribution of code lines and code lengths can be found in Appendix~\ref{appendix:distribution}.
Figure~\ref{fig:line_proportion_v2} shows the percentage of selected code lines. The graph shows that majority of the selected code lines are less than 20\% of the total number of lines of code. 
%This suggests that selecting the correct lines at random is difficult, and requires good understanding of code. \highlight{baseline suggested here might not be a strong one}

% \begin{figure}[h]
% \centering
% \includegraphics[width=0.48\textwidth]{fig/code_percentage_distribution.eps}
% \caption{The distribution of the percentage of selected code lines in code.}
% \label{fig:code_percentage_distribution}
% \end{figure}

% \begin{figure}[h]
% \centering
% \includegraphics[width=0.48\textwidth]{fig/code_density.png}
% \caption{The distribution of the percentage of selected code lines in code.}
% \label{fig:code_span_density}
% \end{figure}

% \begin{table*}[]
% \centering
% \resizebox{\textwidth}{!}{%
% \begin{tabular}{cccccccccc}
% Type & Understanding & Logical Error & Error & Function/Syntax Usage & Algorithm & Task & Reasoning & Explanation & Meaning \\ \hline
% \% lines selected & 0.1104 & 0.2161 & 0.1303 & 0.1185 & 0.1935 & 0.1538 & 0.2188 & 0.4286 & 0.2429 \\ \hline
% \end{tabular}%
% }
% \caption{Percentage of lines selected in code span by type}
% \label{tab:line_percentage}
% \end{table*}

The proportion of \textit{Not applicable} code selections differ by question types, as shown in Table~\ref{tab:question_types}. As TAs ask questions about the implementation details, answering most of them requires looking at the code. On the other hand, students often ask about the approach to the problems and implementation. These questions have less basis on the code and often refer to shorter spans where an error occurs or a function is used. Thus finding the relevant code takes more effort, although answering them requires looking at less code on average.

% The percentage of selected code lines also differs by question types. Questions by TAs have larger percentage of selected code lines, as they often ask about the implemented code. On the other hand, students' questions are less based on the code, and when they do, they often refer to shorter spans where an error occurs or a function is used.

\subsection{Machine Translation Quality}
We have employed 16 workers who are fluent in both Korean and English to check the quality of the machine translation of sampled questions and answers. Each worker checked the quality of 8 question-answer pairs per question type. Each pair was checked by at least two workers. Workers compared the original and the translated texts, and gave scores to four statements on a 5-point Likert scale, with 1 being disagree/bad and 5 being agree/good. The statements were: 1) I can understand the translation, 2) The translation has similar meaning to the original text, 3) The translation contains grammatical and lexical errors, and 4) Overall translation quality. The mean scores between workers for each statement were 4.37, 4.11, 2.06 and 3.92 respectively. The results suggest that the translation was overall in good quality, with high understandability and similar meaning to the original text. The translation contains grammatical or lexical errors, but not to a significant extent.

\section{Experimental Setup}
We select three baseline models, CodeBERT, RoBERTa and XLM-RoBERTa, and test their performance on the type classification and code line selection tasks. CodeBERT model is selected to test the effectiveness of pretraining on NL-PL paired data. Other models based on syntactic structures of code cannot take students' erroneous code as input. RoBERTa and XLM-RoBERTa models are selected to test the performance of NL-based models, for translated and untranslated data respectively. Questions translated to English are provided to the two models pretrained in English, CodeBERT and RoBERTa. XLM-RoBERTa model receives the untranslated questions as input to compare the performance when using the untranslated data. 
We used the default hyperparameters used in CodeBERT for training. 
The tokenizers encode newline token to maintain the code's structure in the tokenized text.
For the code line task, we also test the performance of the naive baseline, which selects the middle 60 lines of code, which showed the best performance among different numbers of lines, as the output.
%The input data was tokenized with RoBERTa tokenizer, which encodes whitespace characters as well, so the tokenized code still maintains the code structure.

Since the token lengths for code in \sysname{} are greater than the limit of the transformer-based models, we preprocess the input to fit within the token length limit. 
We split the code into smaller segments so that the combined length of the split segment and the question is within the limit.
% Each of the combined text of question and split code is given as input to the models for training and inference. 
For type classification, the type with the most number of votes is selected as the final selection.
For code line selection, the model chooses a start and end token position from each segment. The lines between the start and end tokens are given as the output for the segment, and the union of segment outputs is given as the final selection for the question. N/A is given as the output when 1) the end position is before the start position, 2) either the start or the end position is 0 ([CLS] token), or 3) either the start or the end position is out of range.

For the answer retrieval task, we train the DPR by taking the questions as the passages. We use the question with the highest BM25 score in the corpus set as the gold label for the questions in the training set. For testing, the most similar question in the corpus is retrieved using the trained DPR with the new question as the query. The retrieved answer is used as the answer to the new question verbatim.

\section{Results}

We report the mean score from three runs with different seeds for all experiments. The test score is reported on the best-performing epoch out of 10 on the development set.
%The data are split into train, development and test sets in the ratio of 0.6, 0.2 and 0.2 respectively.
%When splitting the data into train, development and test sets, we keep the ratio of question types in each set the same to ensure that there is no unseen data in any of the sets. The test score is evaluated on the best performing epoch out of 10 on the development set.

%We create two datasets for \textit{equal} and \textit{random} type ratios, to understand the effect of question types on the model performance. 
%In the equal setting, each of the question types are equally split into train, validation and test sets. In the random setting, the questions are split into train, validation and test sets without considering the question types.
\subsection{Type Classification}
The results of our baseline models on type classification are shown in Table~\ref{tab:type_classification}.
%There is a slight difference in the scores for the equal and random settings. Generally, the scores for equal setting is higher, suggesting the negative effect of imbalanced data, but the F1 score for RoBERTa on the test set is slightly higher for the random setting. 
The models learn to predict the question types with relatively high accuracy, but there is still a room for improvement.

\begin{table}[h]
\centering
% \small
\resizebox{\columnwidth}{!}{
\begin{tabular}{@{}c|cc|cc|cc@{}}
\toprule
Model    & \multicolumn{2}{c|}{Dev}                             & \multicolumn{2}{c|}{Test}                            & \multicolumn{2}{c}{Q only}                           \\
         & \multicolumn{1}{c|}{Acc}            & F1             & \multicolumn{1}{c|}{Acc}            & F1             & \multicolumn{1}{c|}{Acc}            & F1             \\ \midrule
RoBERTa  & \multicolumn{1}{c|}{\textbf{77.57}} & \textbf{72.31} & \multicolumn{1}{c|}{\textbf{76.65}} & \textbf{71.10} & \multicolumn{1}{c|}{75.74}          & \textbf{69.40} \\
CodeBERT & \multicolumn{1}{c|}{76.20}          & 69.09          & \multicolumn{1}{c|}{75.65}          & 70.13          & \multicolumn{1}{c|}{74.75}          & 67.07          \\ \midrule
XLM-R    & \multicolumn{1}{c|}{72.60}          & 67.88          & \multicolumn{1}{c|}{72.62}          & 66.19          & \multicolumn{1}{c|}{\textbf{76.18}} & 68.68          \\ \bottomrule
\end{tabular}
}
\caption{Type classification task scores for the three baseline models. Q only column shows the test scores with only the question text as the input.}
\label{tab:type_classification}
\end{table}

\begin{table*}[h]
\centering
\resizebox{\textwidth}{!}{%
\begin{tabular}{@{}cc|c|c|c|c|c|c|c|c|c@{}}
\toprule
                          &          & Understanding & Logical & Error & Usage & Algorithm & Task  & Reasoning & Explanation & Meaning \\ \midrule
\multirow{2}{*}{RoBERTa}  & w/ code  & 29.26         & 70.53   & 77.14 & 53.23 & 60.22     & 66.95 & 96.41     & 91.59       & 95.26   \\
                          & w/o code & 21.99         & 70.76   & 76.39 & 50.87 & 67.96     & 68.64 & 95.32     & 88.35       & 94.38   \\ \midrule
\multirow{2}{*}{CodeBERT} & w/ code  & 28.80         & 68.35   & 74.46 & 54.29 & 61.77     & 66.26 & 95.94     & 87.87       & 93.80   \\
                          & w/o code & 13.63         & 67.91   & 74.77 & 44.07 & 59.93     & 65.49 & 95.96     & 87.91       & 93.98   \\ \midrule
\multirow{2}{*}{XLM-R}    & w/ code  & 15.12         & 68.13   & 72.26 & 46.87 & 59.22     & 61.88 & 93.83     & 86.32       & 92.12   \\
                          & w/o code & 9.70          & 71.40   & 75.20 & 53.04 & 61.42     & 67.05 & 97.44     & 88.77       & 94.12   \\ \bottomrule
\end{tabular}%
}
\caption{Class-wise F1 scores on test set for type classification for baseline models}
\label{tab:classwise_f1}
\end{table*}

\begin{table*}[h]
\centering
\resizebox{\textwidth}{!}{%
\begin{tabular}{@{}cc|c|c|c|c|c|c|c|c|c@{}}
\toprule
                          &          & Understanding & Logical & Error & Usage & Algorithm & Task  & Reasoning & Explanation & Meaning \\ \midrule
\multirow{2}{*}{RoBERTa}  & w/ code  & 30.89         & 72.64   & 77.09 & 53.18 & 59.95     & 68.06 & 96.96     & 90.62       & 95.00   \\
                          & w/o code & 33.16         & 72.67   & 77.45 & 53.37 & 59.46     & 66.40 & 96.15     & 90.05       & 94.44   \\ \midrule
\multirow{2}{*}{CodeBERT} & w/ code  & 28.93         & 68.82   & 76.87 & 54.53 & 60.02     & 64.35 & 95.94     & 87.42       & 94.81   \\
                          & w/o code & 30.33         & 65.67   & 72.85 & 54.65 & 60.97     & 67.60 & 96.19     & 88.48       & 94.96   \\ \midrule
\multirow{2}{*}{XLM-R}    & w/ code  & 25.02         & 73.44   & 76.80 & 52.77 & 62.13     & 67.21 & 95.90     & 87.92       & 94.21   \\
                          & w/o code & 28.77         & 69.68   & 77.39 & 54.83 & 59.23     & 67.29 & 96.70     & 88.17       & 93.91   \\ \bottomrule
\end{tabular}%
}
\caption{Class-wise F1 scores on test set for type classification for baseline models trained with augmented data.}
\label{tab:classwise_f1_augmented}
\end{table*}

The class-wise classification F1 scores in Table~\ref{tab:classwise_f1} shows a significant drop for ‘understanding' type when code is not provided. The low number of questions for the understanding type might be the reason, thus we augment the dataset with generated understanding type questions. The common question templates for understanding type questions are extracted, and keywords in the question are randomly replaced with keywords in a randomly chosen code in the dataset. The generated question and the chosen code are given as the input to the models. The question templates are provided in Appendix~\ref{appendix:template}. The class-wise classification F1 scores are reported in Table~\ref{tab:classwise_f1_augmented}. The difference in scores depending on the presence of code is reduced, and overall performance increases. The results suggest that presence of code does not significantly affect the type classification performance. This is expected, as the question type annotation was conducted without providing the code.

\subsection{Code Line Selection}
The results of our baseline models on line selection are shown in Table~\ref{tab:span_selection}. 
%One interesting observation is that the scores on the development set for the random setting is higher, but on the test set, the scores for the equal setting quickly catches up. This could be due to the imbalanced question types in the datasets, and the models in the random setting learns artifacts for the types in the random setting that are not present in the test set. 
We also conduct another set of experiments with questions with N/A line selection removed (Valid Line column). The drop in scores on the code with valid line selections shows that large portion of the scores come from the model correctly identifying N/A selections. 

% \begin{table}[h]
% \centering
% \resizebox{\columnwidth}{!}{%
% \begin{tabular}{c|c|c|c|c|c|c}
% \multirow{2}{*}{Model} & \multicolumn{2}{c|}{Dev} & \multicolumn{2}{c|}{Test} & \multicolumn{2}{c}{Valid Line} \\
%  & EM & F1 & EM & F1 & EM & F1 \\ \hline
% RoBERTa & \textbf{46.62} & \textbf{62.61} & \multicolumn{1}{c|}{\textbf{41.80}} & \textbf{57.57} & \textbf{22.02} & 43.50 \\
% CodeBERT & 42.00 & 57.74 & \multicolumn{1}{c|}{38.95} & 54.06 & 16.42 & 37.12 \\ \hline
% XLM-R & 42.57 & 58.63 & \multicolumn{1}{c|}{39.14} & 55.40 & 21.85 & \textbf{43.90} \\ \hline
% % \textbf{RoBERTa} & \multirow{3}{*}{Random} & \textbf{51.26} & \textbf{64.60} & \multicolumn{1}{c|}{\textbf{42.81}} & \textbf{56.77} \\
% % CodeBERT &  & 45.84 & 60.19 & \multicolumn{1}{c|}{40.06} & 54.37 \\
% % XLM-RoBERTa &  & 47.62 & 62.33 & \multicolumn{1}{c|}{41.04} & 56.51 \\ \hline
% \end{tabular} %
% }
% \caption{The naive and three baseline models' scores on line selection task.}
% \label{tab:span_selection}
% \vspace{-1em}
% \end{table}

\begin{table}[h]
\centering
\resizebox{\columnwidth}{!}{%
\begin{tabular}{@{}c|cc|cc|cc@{}}
\toprule
Model    & \multicolumn{2}{c|}{Dev}                             & \multicolumn{2}{c|}{Test}                            & \multicolumn{2}{c}{Valid Line}                       \\
         & \multicolumn{1}{c|}{EM}             & F1             & \multicolumn{1}{c|}{EM}             & F1             & \multicolumn{1}{c|}{EM}             & F1             \\ \midrule
Naive    & \multicolumn{1}{c|}{1.08}           & 23.97          & \multicolumn{1}{c|}{0.65}           & 21.84          & \multicolumn{1}{c|}{0.90}           & 30.42          \\ \midrule
RoBERTa  & \multicolumn{1}{c|}{\textbf{46.62}} & \textbf{62.61} & \multicolumn{1}{c|}{\textbf{41.80}} & \textbf{57.57} & \multicolumn{1}{c|}{\textbf{22.02}} & 43.50          \\
CodeBERT & \multicolumn{1}{c|}{42.00}          & 57.74          & \multicolumn{1}{c|}{38.95}          & 54.06          & \multicolumn{1}{c|}{16.42}          & 37.12          \\ \midrule
XLM-R    & \multicolumn{1}{c|}{42.57}          & 58.63          & \multicolumn{1}{c|}{39.14}          & 55.40          & \multicolumn{1}{c|}{21.85}          & \textbf{43.90} \\ \bottomrule
\end{tabular}%
}
\caption{The naive and three baseline models' scores on line selection task.}
\label{tab:span_selection}
\end{table}

The naive baseline performance is much worse than the models' performance, which suggests that line selection task is not trivially solved.
The relatively low scores on the tasks for \sysname{} suggest that they are challenging for models built for natural language understanding.
CodeBERT's performance is not superior for the span selection task even though the model was pretrained on code and natural language together. This suggests that CodeBERT's pretraining objective is not appropriate for the~\sysname{} tasks.

% Table~\ref{tab:span_selection_code} shows the performance of the models on the data with actual span selections made. 
% For this dataset, the scores for the equal setting are consistently higher than the scores for the random setting. This shows that when the questions ask for actual code spans, the models trained on the equal setting can provide better selections, rather than choosing to output ``no span''. The result suggests that the models trained on the equal setting can provide more meaning output for use in real course settings.%, irrespective of the numeric score.

% \begin{table}[h]
% \small
% \centering
% \begin{tabular}{c|cc}
% \multirow{2}{*}{Model} & \multicolumn{2}{c}{Test} \\
%  & \multicolumn{1}{c|}{EM} & F1 \\ \hline
% RoBERTa & \multicolumn{1}{c|}{\textbf{22.02}} & 43.50 \\
% CodeBERT & \multicolumn{1}{c|}{16.42} & 37.12 \\ \hline
% XLM-R & \multicolumn{1}{c|}{21.85} & \textbf{43.90} \\ \hline
% % RoBERTa & \multirow{3}{*}{Random} & \multicolumn{1}{c|}{\textbf{19.58}} & 38.73 \\
% % CodeBERT &  & \multicolumn{1}{c|}{15.85} & 35.54 \\
% % XLM-RoBERTa &  & \multicolumn{1}{c|}{19.53} & \textbf{40.70} \\ \hline
% \end{tabular}
% \caption{Span selection task scores for the three baseline models on data without the \textit{Not applicable} code selections.}
% \label{tab:span_selection_code}
% \end{table}

\subsection{Answer Retrieval}
The mean BLEU-1 score that compare the answers for the questions in the test set is 13.07. This shows that a simple retrieval based answering system is not sufficient for answering students' questions. The code provides important context to generate accurate answers, and the answer likely differs even for the same question, depending on the code.

The mean BLEU score for TA's probing questions is 18.48, while that for student-asked questions is 8.91. This suggests that the TAs tend to ask similar questions that have similar answers, while students' questions vary more with largely different answers.

%Additional measures, such as using separate tokenization and word embeddings for source code and natural language are necessary for models to perform well on~\sysname{} tasks, presenting new challenges for code understanding models.}

\subsection{Qualitative Analysis}
In order to better understand the baseline models' behavior, we analyze the output type classifications and line selections for 180 questions, 20 per question type.

For type classification, most of the ‘why’ questions from students are classified as ‘logical error’ or ‘error’ types. These questions are often phrased as “I don’t know why…” or “why something doesn't work”. This leads to relatively high scores for the two error types. 84\% of the ‘why’ questions were classified into the two types. 15 questions were correctly classified.

Keyword matching for line selection task accounts for approximately 54\% of line selections. When a function name or variable name is mentioned in the question, the selected code lines often include the mentioned name. However, this tactic sometimes fools the model into selecting more lines than necessary. This was more frequently observed for \textit{Meaning} and \textit{Function/Syntax Usage} tasks, where 94\% and 75\% of the line selections included the keyword.

\section{Conclusion}
In this paper, we present~\sysname{}, a dataset for code-based question answering in introductory programming course. \sysname{}'s crowdsourced data from a programming course provide rich information that code understanding models need to consider to correctly answer the given questions.
We introduce three tasks for \sysname{}, whose output can help students debug and reduce workloads for the teaching staff. Results from the baseline models indicate that tasks for \sysname{} are challenging for current language understanding models.~\sysname{} promotes further research to better represent and understand source code for code-based question answering. 

As \sysname{} data deliver the full context of the questions, the answer texts in \sysname{} can be used as training and testing data for an answer generation task in the future. Although the generation task is difficult and demands new code representation and processing methods, models that show good performance on it will allow a new level of automation in code-based QA. We hope that \sysname{} will bring research interest in the domain of code understanding for question answering.

\section{Ethical Consideration}
All students and TAs, whose chat logs and code are used to build the dataset, have given permission to use these data for research purposes prior to this research. No disadvantage was given to any student or TAs for not providing their data for this research. The IRB at our university approved the annotation experiments conducted in this research. 

The annotators were compensated appropriately for their participation in the experiments. Compensation was determined to meet the minimum wage requirements.
For the experiment collecting question-answer pairs with question types, the workers were paid \$9 for the first 50 chat logs marked and \$13.50 for every 50 chat logs marked afterwards. It took less than an hour to complete annotations for 50 chat logs on average. 
For the experiment collecting the code lines, the workers were paid \$0.45 for every code annotation made. Workers were able to complete approximately 30 selections in an hour on average.
For the experiment testing the effectiveness of providing relevant code lines on answering the questions, the participants were paid \$13.50 to answer 48 questions by the students. It took approximately an hour for each participant to finish answering all 48 questions.

The authors made their best efforts to anonymize the dataset and remove all personal information such as student ID and phone number from the dataset.

% Entries for the entire Anthology, followed by custom entries
\typeout{}
\bibliography{references}

\begin{thebibliography}{26}
\expandafter\ifx\csname natexlab\endcsname\relax\def\natexlab#1{#1}\fi

\bibitem[{Allamanis et~al.(2018)Allamanis, Brockschmidt, and
  Khademi}]{allamanis2018learning}
Miltiadis Allamanis, Marc Brockschmidt, and Mahmoud Khademi. 2018.
\newblock Learning to represent programs with graphs.
\newblock In \emph{International Conference on Learning Representations}.

\bibitem[{Allamanis and Sutton(2013)}]{allamanis2013and}
Miltiadis Allamanis and Charles Sutton. 2013.
\newblock Why, when, and what: analyzing stack overflow questions by topic,
  type, and code.
\newblock In \emph{2013 10th Working Conference on Mining Software Repositories
  (MSR)}. IEEE.

\bibitem[{Alon et~al.(2018)Alon, Brody, Levy, and Yahav}]{alon2018code2seq}
Uri Alon, Shaked Brody, Omer Levy, and Eran Yahav. 2018.
\newblock code2seq: Generating sequences from structured representations of
  code.
\newblock In \emph{International Conference on Learning Representations}.

\bibitem[{Antol et~al.(2015)Antol, Agrawal, Lu, Mitchell, Batra, Zitnick, and
  Parikh}]{antol2015vqa}
Stanislaw Antol, Aishwarya Agrawal, Jiasen Lu, Margaret Mitchell, Dhruv Batra,
  C~Lawrence Zitnick, and Devi Parikh. 2015.
\newblock Vqa: Visual question answering.
\newblock In \emph{Proceedings of the IEEE international conference on computer
  vision}.

\bibitem[{Brockschmidt et~al.(2018)Brockschmidt, Allamanis, Gaunt, and
  Polozov}]{brockschmidt2018generative}
Marc Brockschmidt, Miltiadis Allamanis, Alexander~L Gaunt, and Oleksandr
  Polozov. 2018.
\newblock Generative code modeling with graphs.
\newblock In \emph{International Conference on Learning Representations}.

\bibitem[{Castelli et~al.(2020)Castelli, Chakravarti, Dana, Ferritto, Florian,
  Franz, Garg, Khandelwal, McCarley, McCawley, Nasr, Pan, Pendus, Pitrelli,
  Pujar, Roukos, Sakrajda, Sil, Uceda-Sosa, Ward, and
  Zhang}]{castelli-etal-2020-techqa}
Vittorio Castelli, Rishav Chakravarti, Saswati Dana, Anthony Ferritto, Radu
  Florian, Martin Franz, Dinesh Garg, Dinesh Khandelwal, Scott McCarley,
  Michael McCawley, Mohamed Nasr, Lin Pan, Cezar Pendus, John Pitrelli, Saurabh
  Pujar, Salim Roukos, Andrzej Sakrajda, Avi Sil, Rosario Uceda-Sosa, Todd
  Ward, and Rong Zhang. 2020.
\newblock The {T}ech{QA} dataset.
\newblock In \emph{Proceedings of the 58th Annual Meeting of the Association
  for Computational Linguistics}, Online. Association for Computational
  Linguistics.

\bibitem[{Clement et~al.(2020)Clement, Drain, Timcheck, Svyatkovskiy, and
  Sundaresan}]{clement2020pymt5}
Colin Clement, Dawn Drain, Jonathan Timcheck, Alexey Svyatkovskiy, and Neel
  Sundaresan. 2020.
\newblock Pymt5: Multi-mode translation of natural language and python code
  with transformers.
\newblock In \emph{Proceedings of the 2020 Conference on Empirical Methods in
  Natural Language Processing (EMNLP)}.

\bibitem[{Cohen(1960)}]{cohen1960coefficient}
Jacob Cohen. 1960.
\newblock A coefficient of agreement for nominal scales.
\newblock \emph{Educational and psychological measurement}, 20(1).

\bibitem[{Conneau et~al.(2020)Conneau, Khandelwal, Goyal, Chaudhary, Wenzek,
  Guzm{\'a}n, Grave, Ott, Zettlemoyer, and Stoyanov}]{conneau2020unsupervised}
Alexis Conneau, Kartikay Khandelwal, Naman Goyal, Vishrav Chaudhary, Guillaume
  Wenzek, Francisco Guzm{\'a}n, {\'E}douard Grave, Myle Ott, Luke Zettlemoyer,
  and Veselin Stoyanov. 2020.
\newblock Unsupervised cross-lingual representation learning at scale.
\newblock In \emph{Proceedings of the 58th Annual Meeting of the Association
  for Computational Linguistics}.

\bibitem[{Feng et~al.(2020)Feng, Guo, Tang, Duan, Feng, Gong, Shou, Qin, Liu,
  Jiang et~al.}]{feng2020codebert}
Zhangyin Feng, Daya Guo, Duyu Tang, Nan Duan, Xiaocheng Feng, Ming Gong, Linjun
  Shou, Bing Qin, Ting Liu, Daxin Jiang, et~al. 2020.
\newblock Codebert: A pre-trained model for programming and natural languages.
\newblock In \emph{Proceedings of the 2020 Conference on Empirical Methods in
  Natural Language Processing: Findings}.

\bibitem[{Guo et~al.(2021)Guo, Ren, Lu, Feng, Tang, LIU, Zhou, Duan,
  Svyatkovskiy, Fu, Tufano, Deng, Clement, Drain, Sundaresan, Yin, Jiang, and
  Zhou}]{guo2021graphcodebert}
Daya Guo, Shuo Ren, Shuai Lu, Zhangyin Feng, Duyu Tang, Shujie LIU, Long Zhou,
  Nan Duan, Alexey Svyatkovskiy, Shengyu Fu, Michele Tufano, Shao~Kun Deng,
  Colin Clement, Dawn Drain, Neel Sundaresan, Jian Yin, Daxin Jiang, and Ming
  Zhou. 2021.
\newblock Graphcode{\{}bert{\}}: Pre-training code representations with data
  flow.
\newblock In \emph{International Conference on Learning Representations}.

\bibitem[{Gupta et~al.(2017)Gupta, Pal, Kanade, and Shevade}]{gupta2017deepfix}
Rahul Gupta, Soham Pal, Aditya Kanade, and Shirish Shevade. 2017.
\newblock Deepfix: Fixing common c language errors by deep learning.
\newblock In \emph{Thirty-First AAAI Conference on Artificial Intelligence}.

\bibitem[{Husain et~al.(2020)Husain, Wu, Gazit, Allamanis, and
  Brockschmidt}]{husain2020codesearchnet}
Hamel Husain, Ho-Hsiang Wu, Tiferet Gazit, Miltiadis Allamanis, and Marc
  Brockschmidt. 2020.
\newblock Codesearchnet challenge: Evaluating the state of semantic code
  search.

\bibitem[{Iyer et~al.(2016)Iyer, Konstas, Cheung, and
  Zettlemoyer}]{iyer2016summarizing}
Srinivasan Iyer, Ioannis Konstas, Alvin Cheung, and Luke Zettlemoyer. 2016.
\newblock Summarizing source code using a neural attention model.
\newblock In \emph{Proceedings of the 54th Annual Meeting of the Association
  for Computational Linguistics (Volume 1: Long Papers)}.

\bibitem[{Kanade et~al.(2020)Kanade, Maniatis, Balakrishnan, and
  Shi}]{kanade2020learning}
Aditya Kanade, Petros Maniatis, Gogul Balakrishnan, and Kensen Shi. 2020.
\newblock Learning and evaluating contextual embedding of source code.
\newblock In \emph{Proceedings of the 37th International Conference on Machine
  Learning}, volume 119 of \emph{Proceedings of Machine Learning Research}.
  PMLR.

\bibitem[{Karpukhin et~al.(2020)Karpukhin, Oguz, Min, Lewis, Wu, Edunov, Chen,
  and Yih}]{karpukhin-etal-2020-dense}
Vladimir Karpukhin, Barlas Oguz, Sewon Min, Patrick Lewis, Ledell Wu, Sergey
  Edunov, Danqi Chen, and Wen-tau Yih. 2020.
\newblock Dense passage retrieval for open-domain question answering.
\newblock In \emph{Proceedings of the 2020 Conference on Empirical Methods in
  Natural Language Processing (EMNLP)}, Online. Association for Computational
  Linguistics.

\bibitem[{Kwiatkowski et~al.(2019)Kwiatkowski, Palomaki, Redfield, Collins,
  Parikh, Alberti, Epstein, Polosukhin, Devlin, Lee
  et~al.}]{kwiatkowski2019natural}
Tom Kwiatkowski, Jennimaria Palomaki, Olivia Redfield, Michael Collins, Ankur
  Parikh, Chris Alberti, Danielle Epstein, Illia Polosukhin, Jacob Devlin,
  Kenton Lee, et~al. 2019.
\newblock Natural questions: A benchmark for question answering research.
\newblock \emph{Transactions of the Association for Computational Linguistics},
  7.

\bibitem[{Liu and Wan(2021)}]{liu2021codeqa}
Chenxiao Liu and Xiaojun Wan. 2021.
\newblock Codeqa: A question answering dataset for source code comprehension.
\newblock In \emph{Findings of the Association for Computational Linguistics:
  EMNLP 2021}.

\bibitem[{Liu et~al.(2019)Liu, Ott, Goyal, Du, Joshi, Chen, Levy, Lewis,
  Zettlemoyer, and Stoyanov}]{liu2019roberta}
Yinhan Liu, Myle Ott, Naman Goyal, Jingfei Du, Mandar Joshi, Danqi Chen, Omer
  Levy, Mike Lewis, Luke Zettlemoyer, and Veselin Stoyanov. 2019.
\newblock Roberta: A robustly optimized bert pretraining approach.
\newblock \emph{arXiv}.

\bibitem[{Rajpurkar et~al.(2016)Rajpurkar, Zhang, Lopyrev, and
  Liang}]{rajpurkar-etal-2016-squad}
Pranav Rajpurkar, Jian Zhang, Konstantin Lopyrev, and Percy Liang. 2016.
\newblock {SQ}u{AD}: 100,000+ questions for machine comprehension of text.
\newblock In \emph{Proceedings of the 2016 Conference on Empirical Methods in
  Natural Language Processing}, Austin, Texas. Association for Computational
  Linguistics.

\bibitem[{Raychev et~al.(2016)Raychev, Bielik, and
  Vechev}]{raychev2016probabilistic}
Veselin Raychev, Pavol Bielik, and Martin Vechev. 2016.
\newblock Probabilistic model for code with decision trees.
\newblock \emph{ACM SIGPLAN Notices}.

\bibitem[{Reddy et~al.(2019)Reddy, Chen, and Manning}]{reddy-etal-2019-coqa}
Siva Reddy, Danqi Chen, and Christopher~D. Manning. 2019.
\newblock {CoQA: A Conversational Question Answering Challenge}.
\newblock \emph{Transactions of the Association for Computational Linguistics},
  7.

\bibitem[{Trischler et~al.(2016)Trischler, Wang, Yuan, Harris, Sordoni,
  Bachman, and Suleman}]{trischler2016newsqa}
Adam Trischler, Tong Wang, Xingdi Yuan, Justin Harris, Alessandro Sordoni,
  Philip Bachman, and Kaheer Suleman. 2016.
\newblock Newsqa: A machine comprehension dataset.
\newblock \emph{arXiv preprint arXiv:1611.09830}.

\bibitem[{Wu et~al.(2016)Wu, Schuster, Chen, Le, Norouzi, Macherey, Krikun,
  Cao, Gao, Macherey, Klingner, Shah, Johnson, Liu, Łukasz Kaiser, Gouws,
  Kato, Kudo, Kazawa, Stevens, Kurian, Patil, Wang, Young, Smith, Riesa,
  Rudnick, Vinyals, Corrado, Hughes, and Dean}]{wu2016gnmt}
Yonghui Wu, Mike Schuster, Zhifeng Chen, Quoc~V. Le, Mohammad Norouzi, Wolfgang
  Macherey, Maxim Krikun, Yuan Cao, Qin Gao, Klaus Macherey, Jeff Klingner,
  Apurva Shah, Melvin Johnson, Xiaobing Liu, Łukasz Kaiser, Stephan Gouws,
  Yoshikiyo Kato, Taku Kudo, Hideto Kazawa, Keith Stevens, George Kurian,
  Nishant Patil, Wei Wang, Cliff Young, Jason Smith, Jason Riesa, Alex Rudnick,
  Oriol Vinyals, Greg Corrado, Macduff Hughes, and Jeffrey Dean. 2016.
\newblock Google's neural machine translation system: Bridging the gap between
  human and machine translation.
\newblock \emph{CoRR}.

\bibitem[{Yang et~al.(2018)Yang, Qi, Zhang, Bengio, Cohen, Salakhutdinov, and
  Manning}]{yang2018hotpotqa}
Zhilin Yang, Peng Qi, Saizheng Zhang, Yoshua Bengio, William Cohen, Ruslan
  Salakhutdinov, and Christopher~D Manning. 2018.
\newblock Hotpotqa: A dataset for diverse, explainable multi-hop question
  answering.
\newblock In \emph{Proceedings of the 2018 Conference on Empirical Methods in
  Natural Language Processing}.

\bibitem[{Yasunaga and Liang(2020)}]{yasunaga2020graph}
Michihiro Yasunaga and Percy Liang. 2020.
\newblock Graph-based, self-supervised program repair from diagnostic feedback.
\newblock In \emph{International Conference on Machine Learning}, pages
  10799--10808. PMLR.

\end{thebibliography}

\bibliographystyle{acl_natbib}

\newpage
\appendix
\onecolumn
\section*{Appendix}
\section{Example data in \sysname{}}
\label{appendix:example}
\begin{figure*}[h]
\centering
\includegraphics[width=\textwidth]{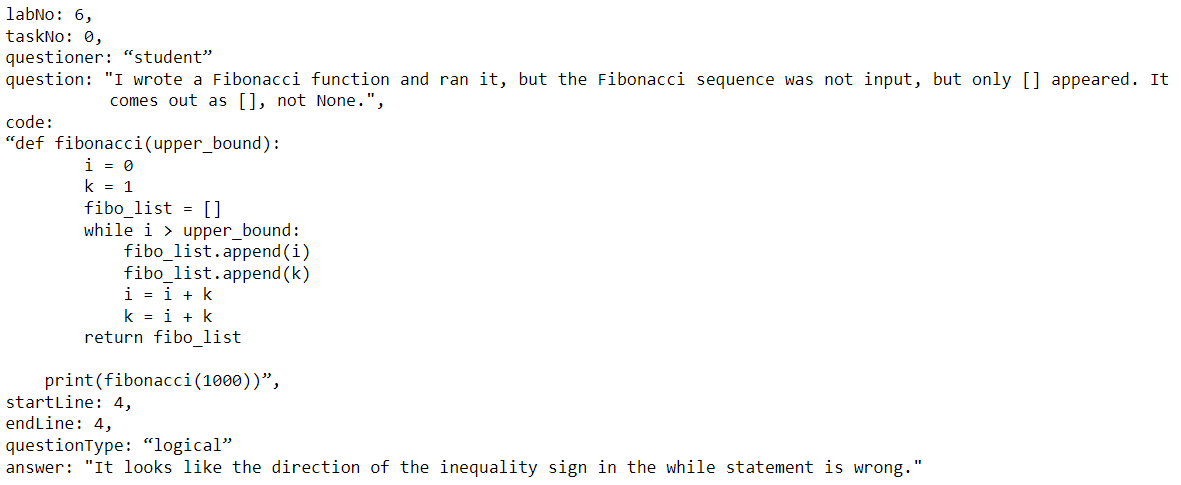}
\caption{An example of the data in \sysname{}. Note that taskNo, startLine and endLine variables count from 0. The code is prettified for readability.}
\label{fig:example_data}
\end{figure*}

We present an example of a question in the \sysname{} dataset in Figure~\ref{fig:example_data}.

\clearpage
\twocolumn
\section{Distribution of Question, Answer and Code Lengths}
\label{appendix:distribution}

\begin{figure}[h!]
\centering
\includegraphics[width=0.41\textwidth]{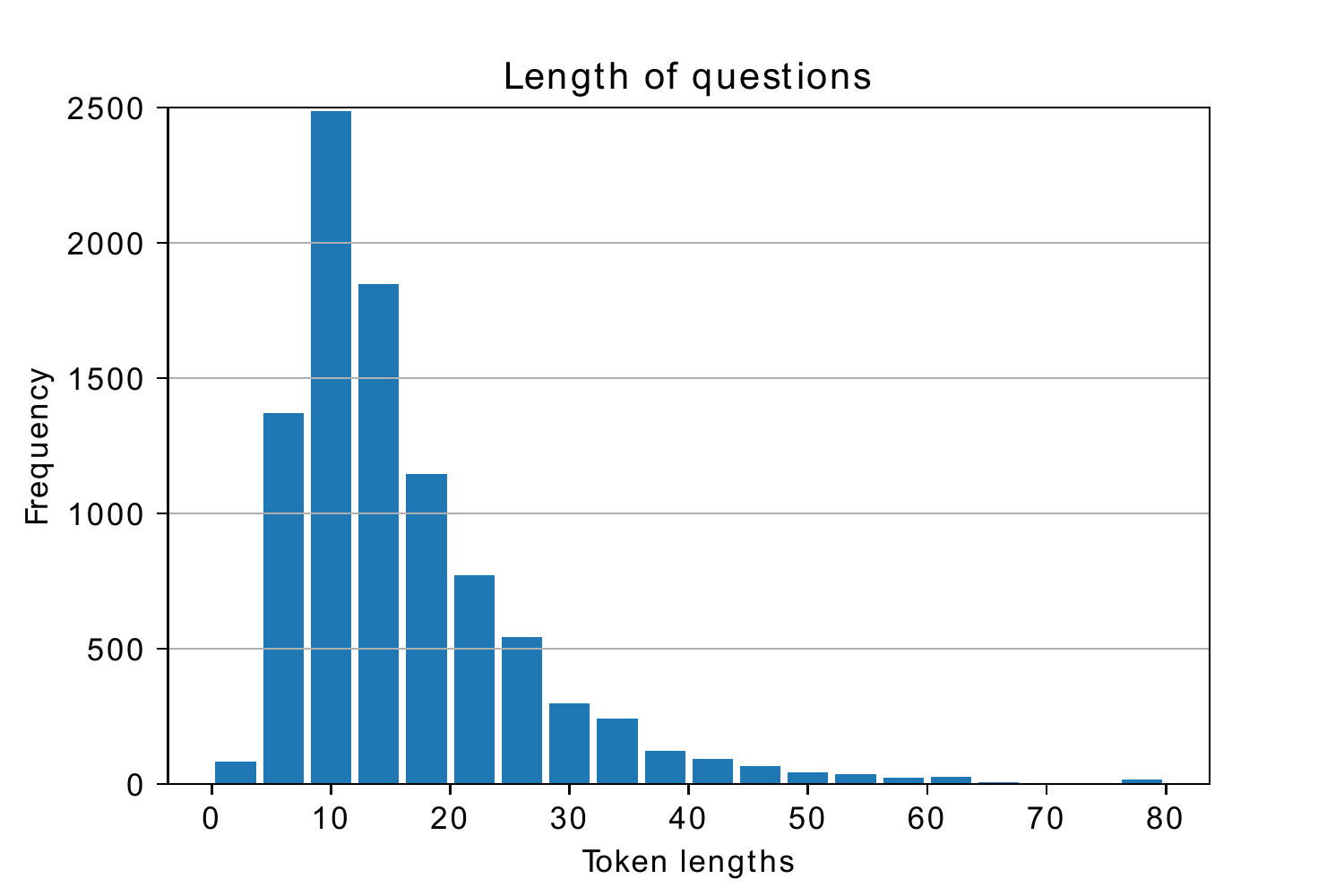}
\caption{The distribution of question lengths translated to English in number of white space separated tokens. The last bin contains all questions longer than 80 tokens.}
\label{fig:question_length_distribution}
\end{figure}

\begin{figure}[h!]
\centering
\includegraphics[width=0.41\textwidth]{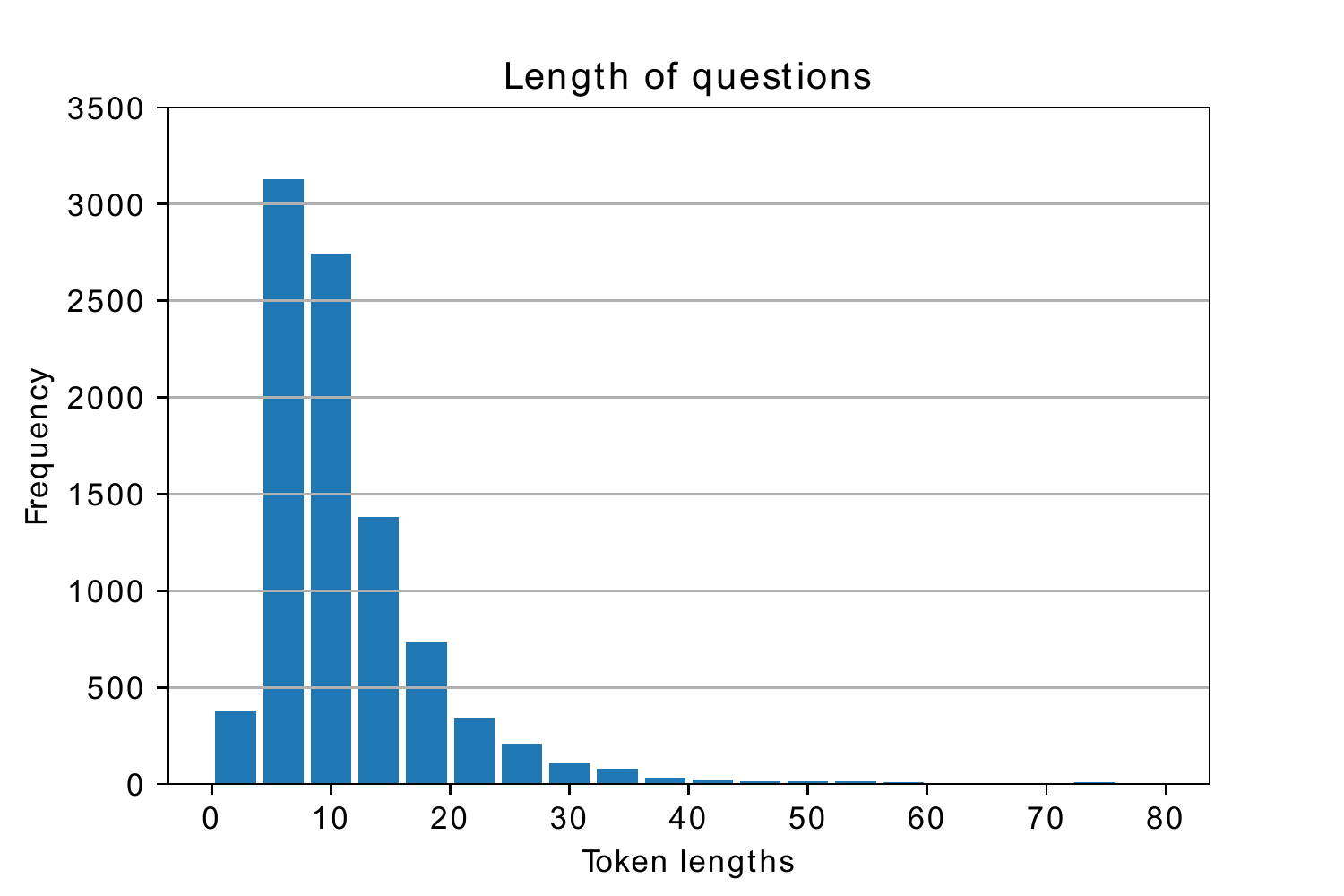}
\caption{The distribution of original question lengths in number of white space separated tokens. The last bin contains all questions longer than 80 tokens.}
\label{fig:question_length_distribution_orig}
\end{figure}

Figures \ref{fig:question_length_distribution} and \ref{fig:question_length_distribution_orig} show the distribution of question lengths for questions translated to English and original questions respectively.

\begin{figure}[h!]
\centering
\includegraphics[width=0.41\textwidth]{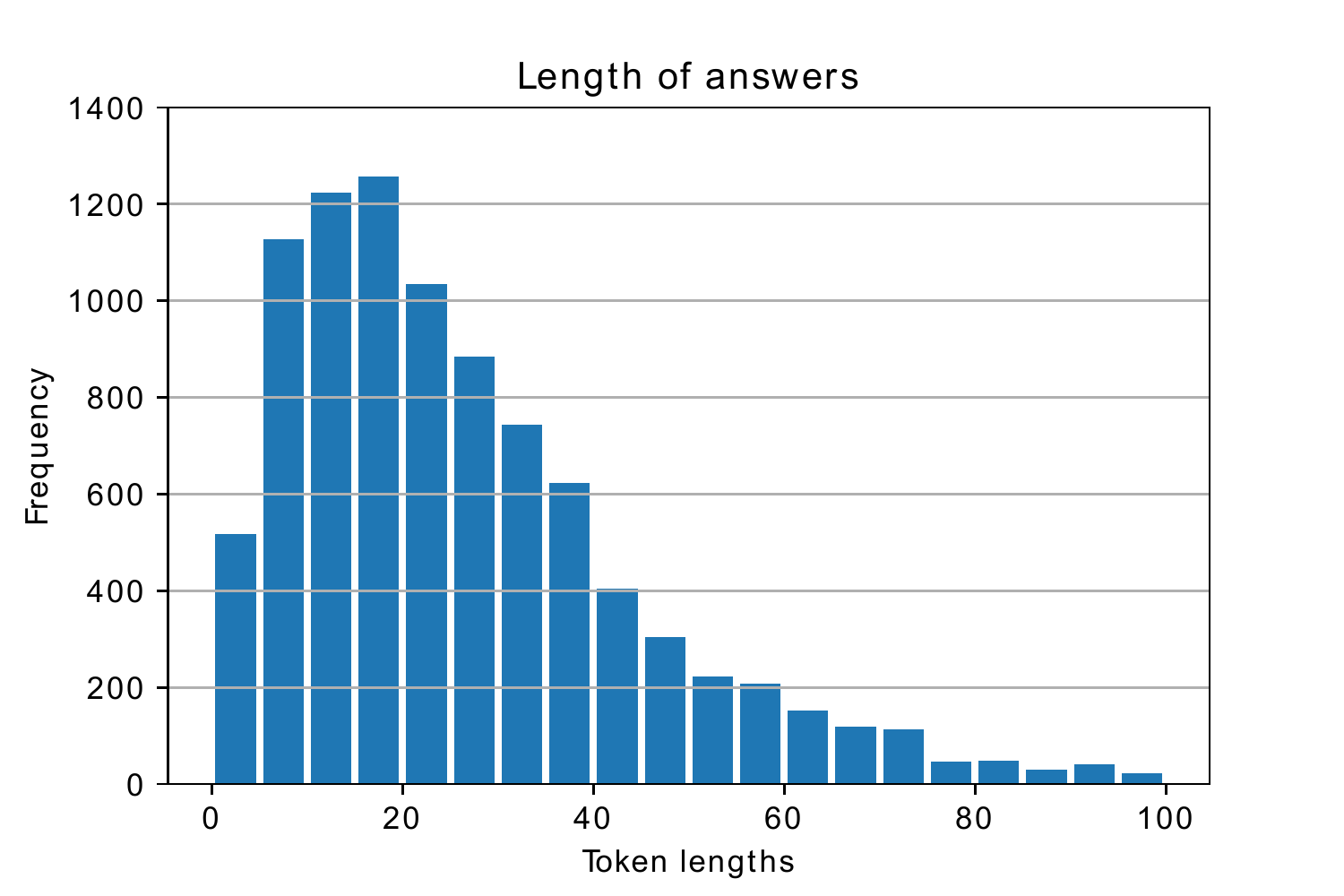}
\caption{The distribution of answer lengths in number of white space separated tokens. The last bin contains all answers longer than 100 tokens.}
\label{fig:answer_length_distribution}
\end{figure}

\begin{figure}[h!]
\centering
\includegraphics[width=0.41\textwidth]{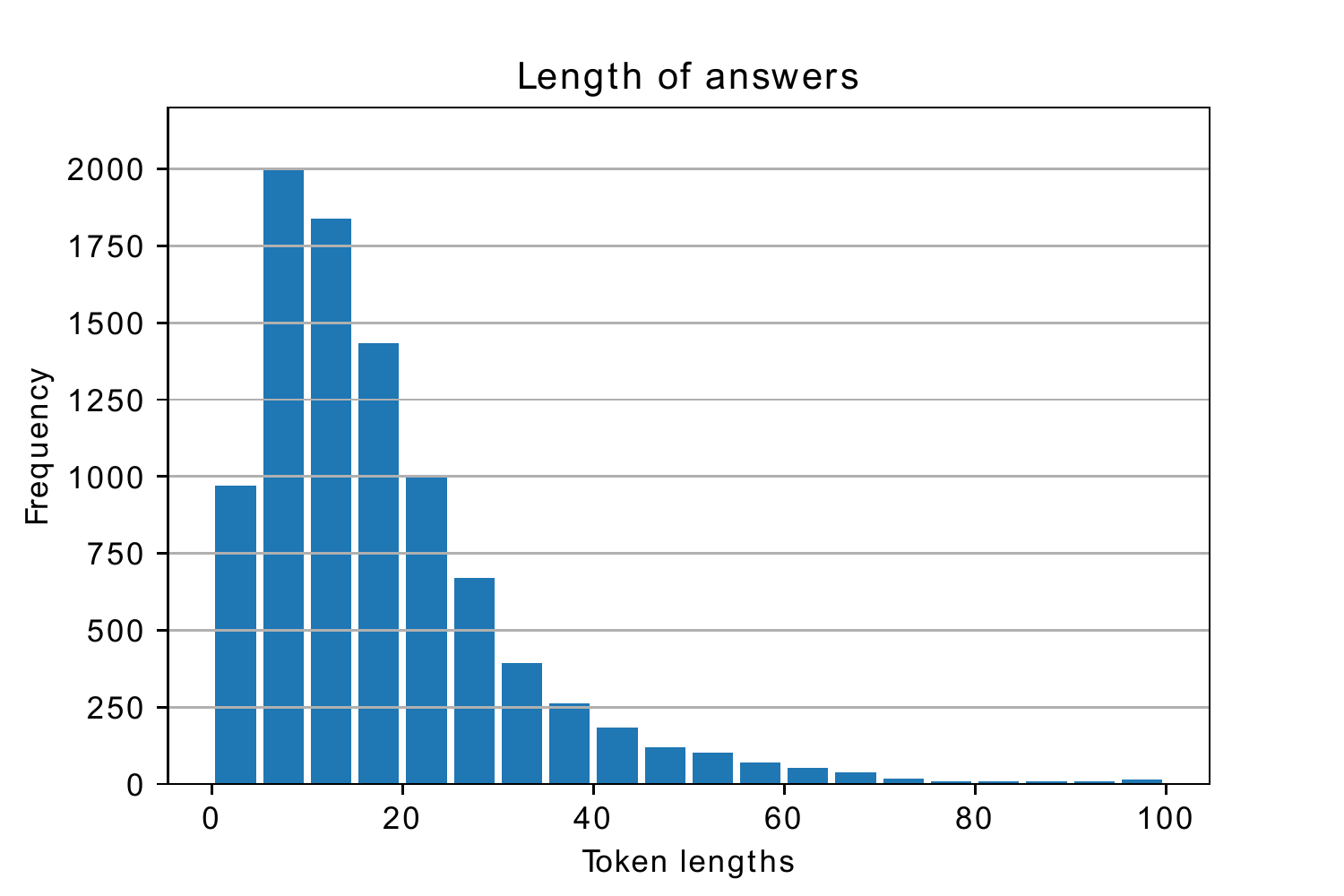}
\caption{The distribution of original answer lengths in number of white space separated tokens. The last bin contains all answers longer than 100 tokens.}
\label{fig:answer_length_distribution_orig}
\end{figure}

Figures \ref{fig:answer_length_distribution} and \ref{fig:answer_length_distribution_orig} show the distribution of answer lengths for answers translated to English and original answers respectively.

\begin{figure}[h!]
\centering
\includegraphics[width=0.41\textwidth]{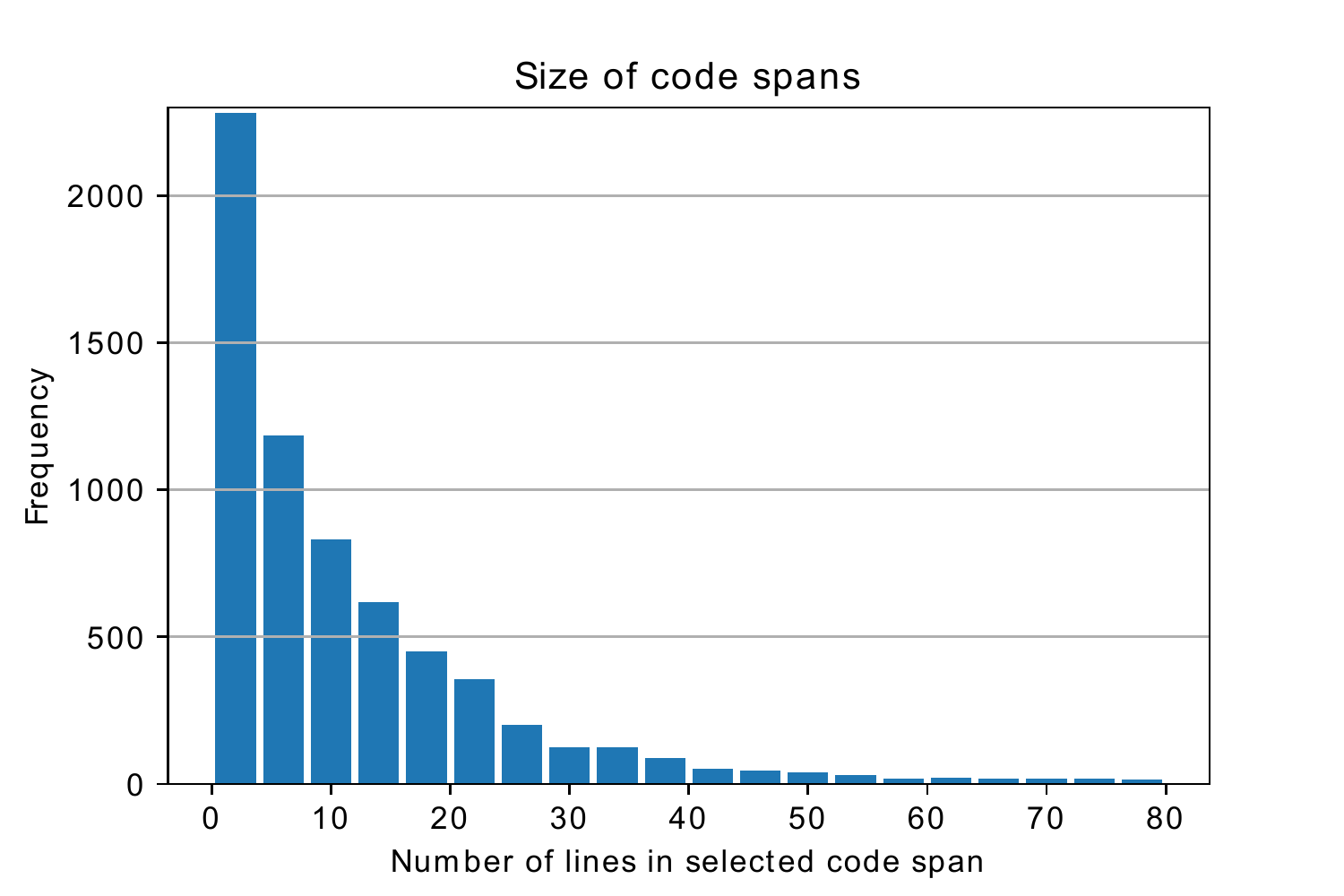}
\caption{The distribution of number of lines selected in code spans. The last bin contains all selections with more than 80 lines of code.}
\label{fig:code_span_distribution}
\end{figure}

\begin{figure}[h!]
\centering
\includegraphics[width=0.41\textwidth]{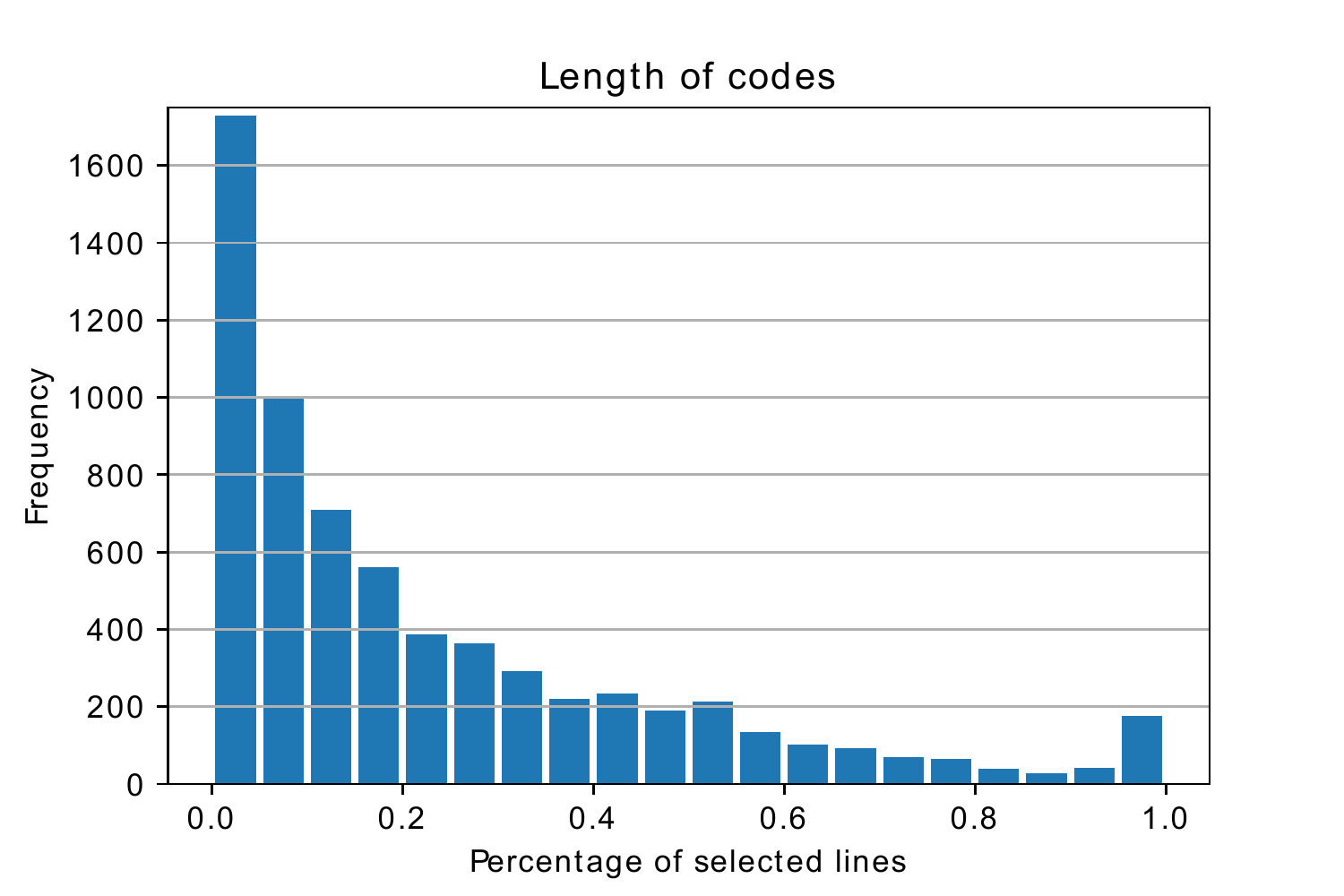}
\caption{The distribution of the percentage of selected code lines in code.}
\label{fig:code_percentage_distribution}
\end{figure}

Figure \ref{fig:code_span_distribution} shows the distribution of number of lines in the selected code spans. Figure \ref{fig:code_percentage_distribution} shows the distribution of proportions of the code lines that is included in the selected code span.

\clearpage
\onecolumn

\section{Question Templates for Understanding Type Questions}
\label{appendix:template}
The templates used for augmenting~\sysname{} dataset with understanding type questions are presented in Table~\ref{tab:understanding_templates}. The keywords variable, function, and snippet are extracted from the randomly chosen code in the dataset. Variable is a random token, function is a random function name, and snippet is a random line of code. In the last template, one of the words list, dictionary, variable, function is chosen randomly to complete the template.

\begin{table}[h]
\begin{tabularx}{\textwidth}{l}
Template format                                                         \\ \hline
What does {[}variable, function{]} mean?                                \\
What does {[}variable, function{]} refer to?                            \\
What's the meaning of {[}variable, function{]}                          \\
What does {[}function{]} do?                                            \\
Can you explain what {[}function{]} does?                               \\
Can you describe what {[}function{]} is doing?                          \\
How do I use {[}function{]}?                                            \\
How to use {[}function{]}?                                              \\
I don't understand {[}snippet{]}.                                       \\
What is {[}function{]}?                                                 \\
Should I use {[}function, snippet{]}?                                   \\
Why do you do {[}snippet{]}?                                            \\
Is {[}variable, function{]} a \{list, dictionary, variable, function\}?
\end{tabularx}
\caption{Templates used for the question augmentation for Understanding type questions. The keywords in square brackets are chosen from a randomly chosen code in the dataset. The words in curly brackets are randomly chosen.}
\label{tab:understanding_templates}
\end{table}

\section{Experiment Details}
\label{appendix:experiment_details}
We ran the experiments for RoBERTa-base, CodeBERT-base and XLM-RoBERTa model on 4 Quadro RTX 8000 GPUs. We ran 10 epochs for fine-tuning the models. All of these models were released with MIT License, and our use is consistent with the license.

For all models, we used the batch size of 32 for training, evaluating and testing. 

The average runtime for each epoch for RoBERTa-base and CodeBERT-base models is approximately 1 hour for training, and 1 minute for evaluating and testing. For XLM-RoBERTa-base model, the average runtime for each epoch is approximately 3.3 minutes hour for training and 0.5 minute for evaluating and testing.

The number of parameters for RoBERTa-base, CodeBERT-base and XLM-RoBERTa models are 125M, 125M and 270M respectively.
\clearpage

\section{Annotation Interface}
\label{appendix:annotation}
\begin{figure*}[h]
\centering
\includegraphics[width=0.97\textwidth]{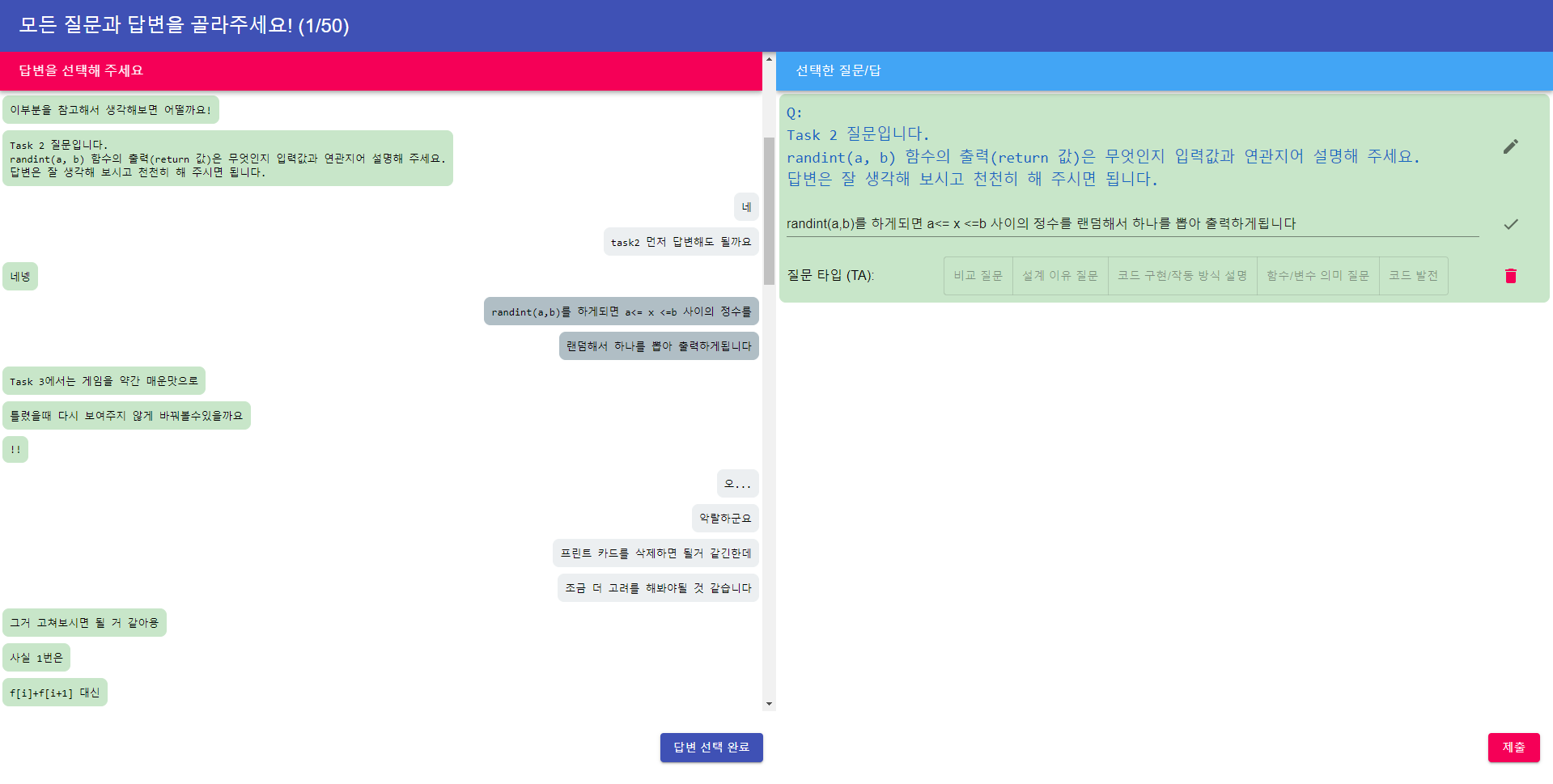}
\caption{The annotation interface for question, answer and type selection. On the left, the chat log is presented. On the right, annotators can modify the question and answer texts and select the question type.}
\label{fig:qa_annotation}
\end{figure*}

We present the annotation interface used to collect the question, answer, and question type in Figure~\ref{fig:qa_annotation}. Annotators can choose the messages corresponding to the question or answer text, and modify the texts in the interface. Annotators also select a question type for every question.

\begin{figure*}[h]
\centering
\includegraphics[width=0.97\textwidth]{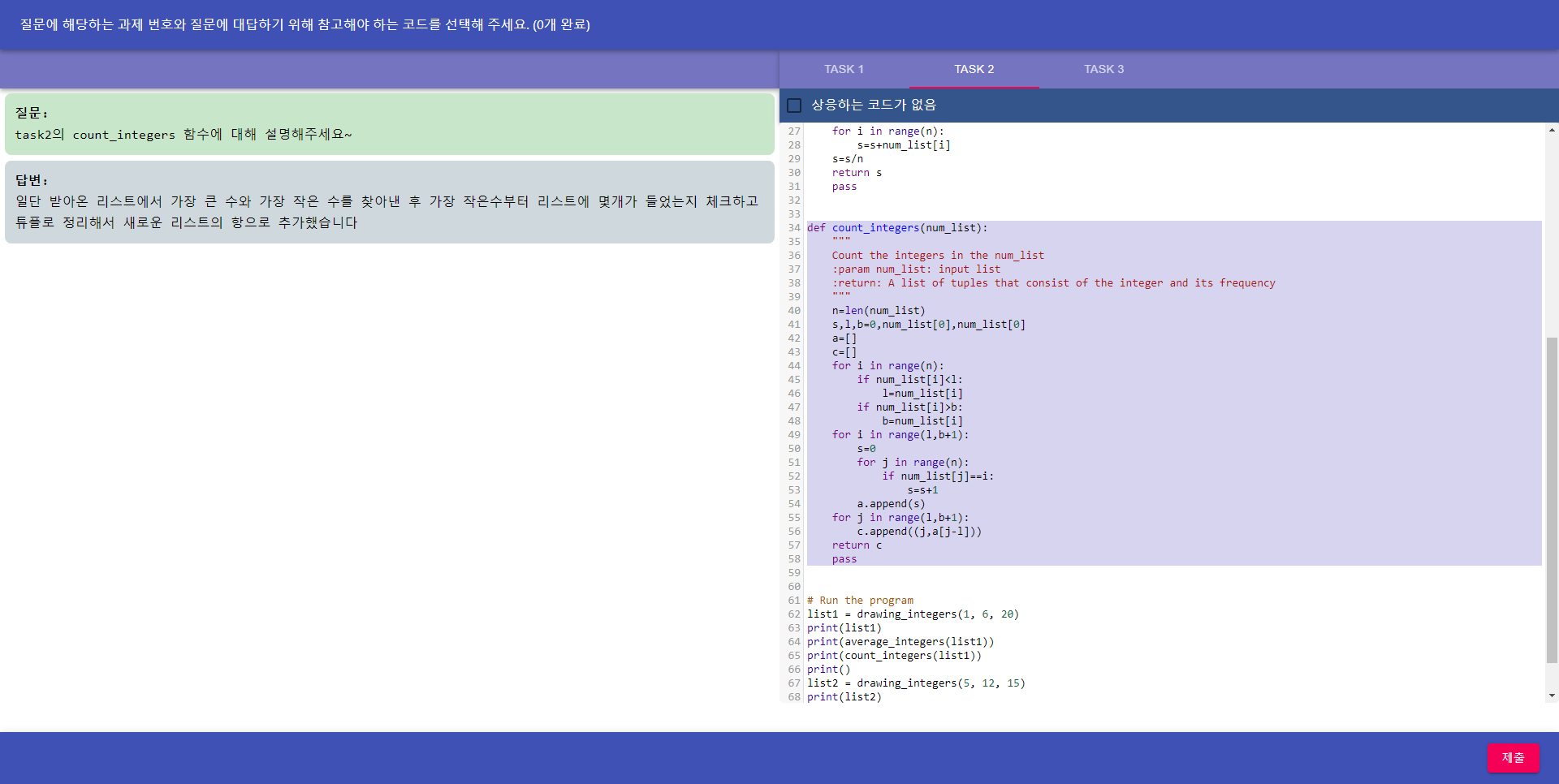}
\caption{The annotation interface for code line selection. On the left, the question and answer texts are presented. On the right, annotators select the correct task for the given question and answer, and select the code lines that provide information to answer the question.}
\label{fig:code_annotation}
\end{figure*}

We present the annotation interface used to collect the code and the code span in Figure~\ref{fig:code_annotation}. Annotators choose the code for the question given, and select code spans with a code line as a unit.

The full-text instructions for QA annotation can be found in \href{https://docs.google.com/document/d/1Q6HJc2K9j-4btK3PdRVjLgzJ11twe8quqzSX6YJmnWU/edit?usp=sharing}{this link}.
The instructions for code line annotaion can be found in \href{https://docs.google.com/document/d/1s-X7RHjZMQd2iqP5RPSzjrvBEy267NfPqYTHPUlxI60/edit?usp=sharing}{this link}.

% \section{Line Selection Results on the Same Code with Changing Questions}

% \begin{figure}[h]
% \centering
% \includegraphics[width=0.75\linewidth]{fig/code_selected.PNG}
% \caption{Line selection results when question changes. The red box shows the selection for the question ``I keep getting an error while True:.'', and the blue box shows the selection for the question ``The code does not stop.''}
% \label{fig:qual_code_1}
% \end{figure}

\end{document}